\documentclass[letterpaper]{article} % DO NOT CHANGE THIS
\usepackage{aaai2027}  % DO NOT CHANGE THIS
\usepackage[hyphens]{url}  % DO NOT CHANGE THIS
\usepackage{graphicx} % DO NOT CHANGE THIS
\def\UrlFont{\rm}  % DO NOT CHANGE THIS
\usepackage{natbib}  % DO NOT CHANGE THIS AND DO NOT ADD ANY OPTIONS TO IT
\usepackage{caption} % DO NOT CHANGE THIS AND DO NOT ADD ANY OPTIONS TO IT
\usepackage{algorithm}
\usepackage{algorithmic}
\usepackage{newfloat}
\usepackage{multirow}
\usepackage{amsmath}
\usepackage{amssymb}
\usepackage{pifont}

\usepackage{newfloat}
\usepackage{listings}
\DeclareCaptionStyle{ruled}{labelfont=normalfont,labelsep=colon,strut=off} % DO NOT CHANGE THIS
\floatstyle{ruled}
\newfloat{listing}{tb}{lst}{}
\floatname{listing}{Listing}

\usepackage{booktabs}

\title{InfoDense: Density-Aware Regional Decisive Replay for Memory-Efficient Incremental Face Forgery Detection}
\author{
Jikang Cheng$^{2*}$, Hao Shen$^{1*}$, Xueyi Zhang$^{3}$, Guangcheng Wang$^{4}$ \\
Zhongyuan Wang$^{5}$, Renye Yan$^{2\dagger}$, Baojin Huang$^{1\dagger}$ \\[0.2cm]
$^{1}$Huazhong Agricultural University, $^{2}$Peking University \\
$^{3}$National University of Singapore, $^{4}$Nantong University \\
$^{5}$Wuhan University \\[0.2cm]
\small $^{*}$These authors contributed equally to this work \\
\small $^{\dagger}$Correspondence to: victory@stu.pku.edu.cn, huangbaojin@mail.hzau.edu.cn
}
\nocopyright
\affiliations{
}

\begin{document}

\maketitle

\begin{abstract}
  The rapid evolution of face forgery techniques has introduced an increasing variety of manipulations. Incremental Face Forgery Detection (IFFD), which incrementally adds new forgery data to fine-tune previously trained models, has emerged as a promising approach to handle evolving forgery threats. However, conventional replay-based IFFD methods suffer from catastrophic forgetting. Storing full historical images under limited memory often either fails to preserve subtle forgery cues or introduces domain bias, reducing the model’s ability to learn intrinsic and transferable manipulation characteristics. In this paper, we propose a Density-Aware Regional Decisive replay strategy, termed \textbf{InfoDense}, to address these challenges. InfoDense prioritizes artifact-dense and forgery-critical regions, significantly reducing storage requirements while maintaining high-fidelity forgery evidence. We first introduce \textbf{InfoDense Cut} to localize decisive patches using CLIP-based embeddings. Then, \textbf{InfoDense Select} ranks candidate segments by combining latent-space representativeness and decisive patch counts, ensuring both diversity and information density in the replay buffer. Finally, \textbf{InfoDense Fuse} reconstructs unbiased training inputs by adaptively merging stored segments with current-task samples, enhancing knowledge retention and generalization. Extensive experiments on challenging incremental deepfake benchmarks demonstrate that InfoDense effectively mitigates catastrophic forgetting while improving cross-domain generalization.
\end{abstract}

\section{Introduction}
The rapid advancement of generative technologies has led to increasingly diverse and realistic forgery faces, posing serious threats to multimedia security and social trust. One line of research seeks to address this challenge by training forgery detection models with strong generalization ability under limited data conditions, so as to improve robustness to unseen forgeries. However, generalization-oriented approaches \cite{sun2022dual,cao2022end,huang2023implicit,yan2024transcending,cheng2024can,cheng2025ed} can only partially handle unknown forgery patterns and still struggle to cope with the continuous evolution of forgery techniques. To this end, Incremental Face Forgery Detection (IFFD) has recently attracted growing attention, aiming to progressively learn new forgery types while maintaining reliable detection performance on previously learned ones.

IFFD methods typically mitigate catastrophic forgetting by replaying representative samples during the learning of new data. Common strategies include central sample replay \cite{pan2023dfil}, hard sample replay \cite{pan2023dfil}, adversarial perturbation–based replay \cite{sun2025continual}, mixed prototypes \cite{tian2024dynamic}, and sparse uniform replay \cite{cheng2025stacking}. However, these replay-based approaches inevitably require storing raw historical samples, resulting in non-negligible memory overhead and potential privacy risks. Please refer to Appendix Section~\ref{related work} for detailed illustrations of \textbf{\textit{Related Work}}. Notably, face images exhibit strong structural regularities, such as bilateral symmetry and the spatial concentration of discriminative cues. For deepfake detection, the information required for effective replay mainly concerns identity-related features, forgery artifacts, and domain-specific patterns. These cues are typically localized and redundantly distributed across symmetric facial regions rather than uniformly over the entire face. Hence, storing the full face image may be unnecessary to preserve replay knowledge, while selectively retaining informative regions can sufficiently maintain the critical information needed to mitigate forgetting in IFFD.

\begin{figure}[t]
    \centering
    \includegraphics[width=1\linewidth]{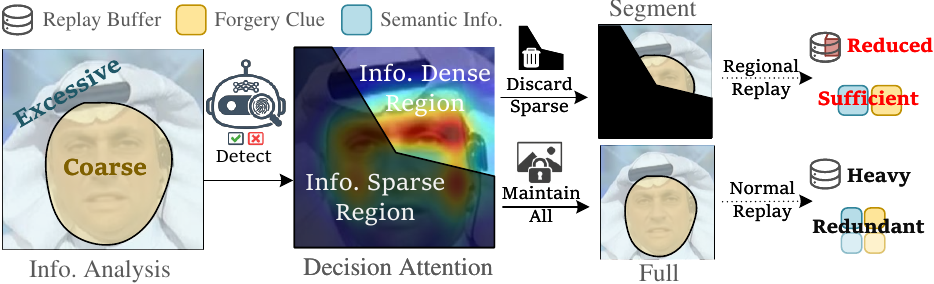}
    \caption{Intuitive impression of our motivation: replaying InfoDense segment instead of replaying redundant full faces.}
    \label{fig:intro}
\end{figure}

\begin{figure}[t]
    \centering
    \includegraphics[width=1\linewidth]{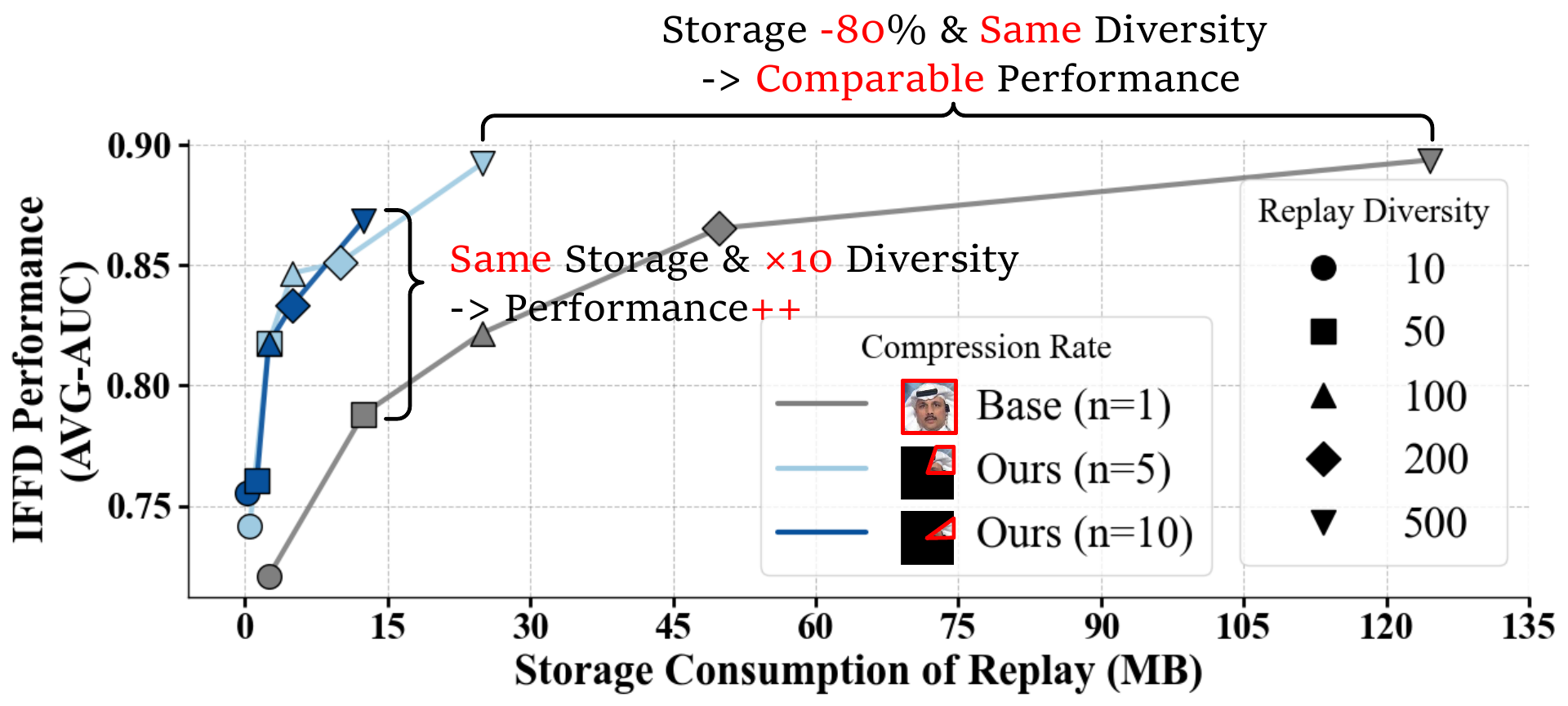}
    \caption{Performance impression on compressing redundant facial information. The results are obtained with Protocol 1 (see Sec.~\ref{sec:exp_set}). \textbf{n} in Compression Rate denotes $\frac{1}{\text{n}}$ segment of one intact image is used for replay. Replay Diversity is represented by the number of image samples stored in the replay set. With significantly reduced storage, two different versions of our approach achieve comparable performance against the baseline, indicating the redundancy of using intact face images as the replay samples.}
    \label{fig:illustration}
\end{figure}

Motivated by the above observations, we aim to design a \textbf{regional decisive replay} strategy that preserves essential forgery knowledge while significantly reducing storage cost. As shown in Fig. \ref{fig:intro}, we model and visualize the decisive attention of samples during detector training. According to the resulting attention distribution, forgery face images can be conceptually divided into \textbf{information-dense} and \textbf{information-sparse} regions. The \textbf{former} is sufficient to capture critical domain characteristics and \textbf{forgery traces}, while the \textbf{latter} mainly contains \textbf{redundant content}. Furthermore, the empirical results in Fig. \ref{fig:illustration} validate the substantial potential of compressing redundant facial information. Even when only a fraction of each image is retained, the detection performance remains highly competitive, indicating that intact facial images contain considerable redundancy for replay. Based on this insight, we preferentially retain information-dense regions during replay, \textbf{enabling effective knowledge preservation with significantly reduced memory cost}. As a result, more historical samples can be stored under the same budget, alleviating catastrophic forgetting. Meanwhile, concentrating replay on compact yet decisive regions encourages the detector to focus on intrinsic forgery cues, thereby improving generalization to unseen manipulations.

In this paper, we propose a Density-Aware Regional Decisive replay strategy, termed InfoDense, for memory-efficient incremental deepfake detection. The goal is to preserve forgery-critical information under limited replay capacity. We first introduce \textbf{InfoDense Cut} to estimate the regional density of decisive forgery cues. Patch-level image tokens are extracted using a CLIP-based encoder. Each token is assigned a forgery-decisive score based on its similarity to forgery-representative text embeddings. The top-$K$ patches are selected as Decisive Patches to localize the most discriminative regions. We then perform \textbf{InfoDense Select} to rank candidate segments. The ranking is based on a weighted combination of latent-space centroid distance and decisive patch count. The top-$m$ segments are stored in the replay buffer to ensure both representativeness and information density. Finally, we apply \textbf{InfoDense Fuse} to adaptively combine stored segments with current-task samples. This process reconstructs complete training inputs and enables effective and unbiased replay. It also enriches fine-grained forgery evidence and mitigates catastrophic forgetting during incremental learning. Our contributions can be summarized as follows:
(1) By exploiting the inherent redundancy in facial information, InfoDense significantly reduces storage requirements. Under the same memory budget, it enables a multiplicative increase in replay diversity, thereby effectively alleviating catastrophic forgetting.
(2) InfoDense distills core forgery-decisive cues and reuses them across multiple tasks, encouraging the model to capture intrinsic forgery characteristics rather than task-specific artifacts, thus improving generalization ability.
(3) Since the replay buffer stores only compact decisive regions instead of complete facial images, InfoDense naturally mitigates potential privacy risks associated with data replay, making it more suitable for practical deployment.
% \textcolor{red}{Yry:There are too many lines with only a few letters or words occupying an entire line.}

\begin{figure*}[t]
    \centering
    \includegraphics[width=1\linewidth]{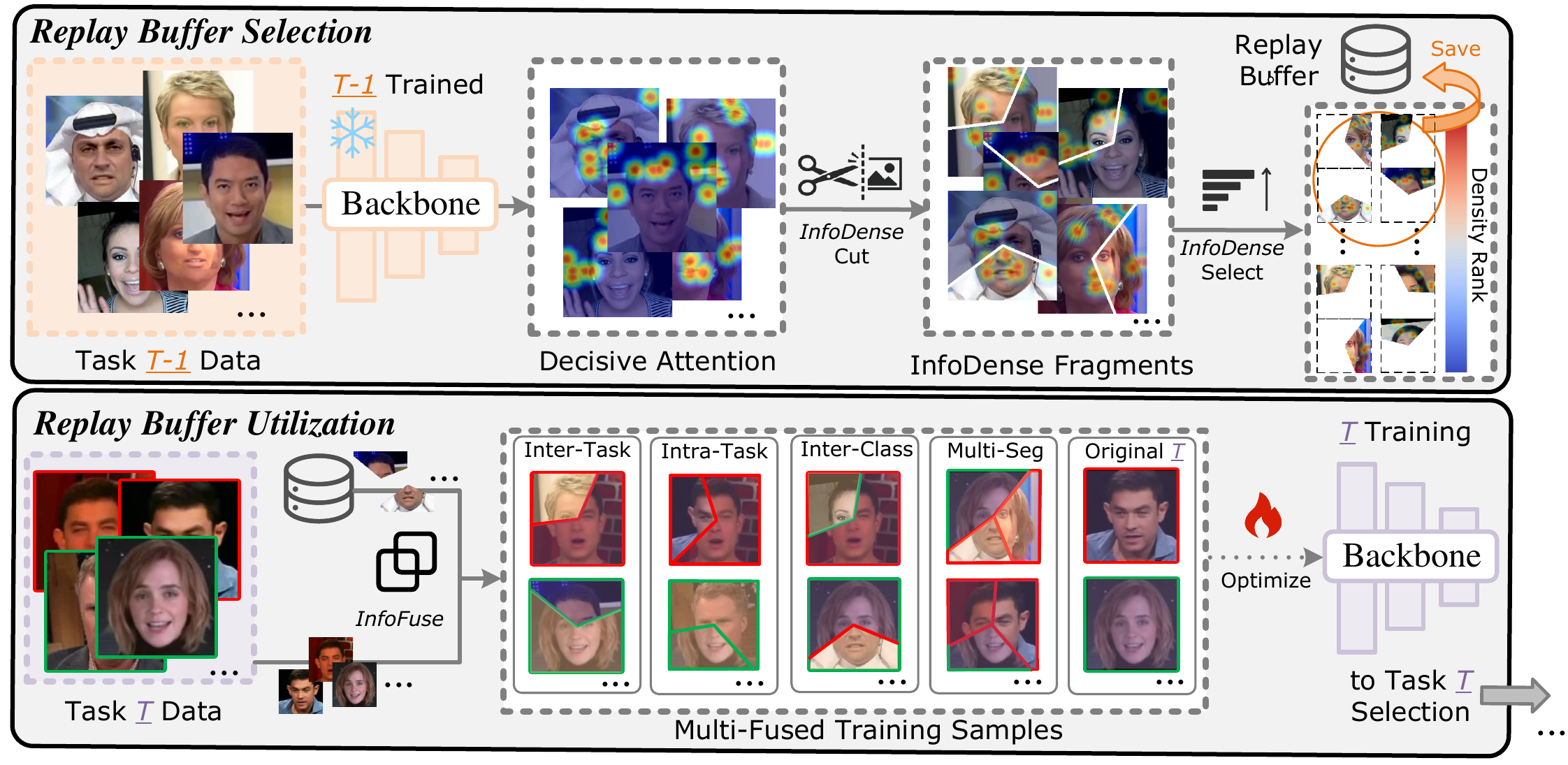}
    \caption{The overall pipeline of the proposed InfoDense. We extract informative regional segments via a cut-and-select strategy. These segments are then used in replay learning to mitigate catastrophic forgetting and enhance generalization.}
    \label{fig:pipeline}
\end{figure*}
\section{Proposed Method}
To compress the redundant information in the facial data, thus improving the diversity and information density under the limit of the same storage consumption, we propose a unified framework termed InfoDense that can cut, select, and utilize the regional segments with dense information for replay. The overall pipeline can be found in Fig.~\ref{fig:pipeline}.
\subsection{Rationale Behind Compressing Facial Information}
Facial images contain substantial structural redundancy due to the strong spatial regularity and identity consistency of human faces. A large portion of a face, including smooth skin regions such as the cheeks, forehead, and jaw area, exhibits a homogeneous appearance that remains stable across frames, poses, and expressions. These regions primarily encode identity-related information and contribute little to forgery detection. In contrast, deepfake detectors mainly rely on subtle local inconsistencies introduced during synthesis and blending, which are typically concentrated around semantically sensitive regions such as the eyes, mouth, nose boundary, and facial contour. As a result, the majority of facial pixels provide highly correlated and repetitive signals from the perspective of the detection objective.

Moreover, recent deepfake generation methods aim to preserve global facial structure and identity while minimizing perceptual artifacts, which further increases redundancy in non-critical regions. The discriminative evidence is therefore sparse relative to the full image, and storing entire facial images for replay retains large amounts of non-informative content. By compressing facial information and preserving only forgery-decisive regions, the replay strategy focuses on task-relevant evidence while discarding redundant identity-consistent signals. This selective retention reduces memory consumption and enables more diverse replay samples under a fixed storage budget, which is particularly beneficial for incremental deepfake detection.
\subsection{Replay Selection}
\subsubsection{Regional Density of Decisive Information.}
First, we manage to expose the region in a single face that is the most decisive for the judgment of the detectors. At present, a variety of well-established techniques have been developed to obtain the decision-related attention regions of a model (either CNN-based~\cite{selvaraju2017gradcam,chattopadhay2018gradpp} or ViT-based~\cite{bousselham2025legrad,chefer2021transformergrad}) with respect to a given input image. Building upon these approaches, we incorporate forged information and propose to leverage the text-image alignment capacity of CLIP to highlight the forgery-related similarity of different patches. Specifically, given an input image $\mathbf{x}^{t-1} \in \mathbb{I}^{t-1}$, where $\mathbb{I}^{t-1}$ denotes the $t-1$ task, we can have a CLIP-based detector $f(\cdot,\theta^{t-1})$ with $t-1$ trained parameters $\theta^{t-1}$. We can then obtain a series of image tokens using $t-1$ encoder ($\mathit{E}(\cdot)$) as $\mathit{E}(\mathbf{x}^{t-1})=\{\mathbf{z}_{cls},\mathbf{z}_{(1,1)},...,\mathbf{z}_{(n,n)}\}$. Subsequently, to evaluate the forgery-decisive score ($\delta$) of each token, we compare the similarity of each token with the forgery-representative token ($\mathbf{z}_{f}$), which is obtained by encoding forgery-representative prompts via the text encoder. Since both text and image tokens are pre-trained to be mapped to the consistent latent space, they can be directly leveraged to compare similarity. Formally, the forgery decisive score $\delta$ of the patch at location $(i,j)$ is computed as:
\begin{equation}
    \delta_{(i,j)}=\text{Sim}(\mathbf{z}_{(i,j)},\mathbf{z}_{f}),
\end{equation}
where $\text{Sim}(\cdot)$ calculates the cosine similarity. Appendix~\ref{sec:supp-clip} provides an extensive analysis and empirical support for adopting this CLIP-based attention algorithm, a robust approach widely validated in prior work~\cite{ForAda,SimplicityAIGI}. Finally, we sort the patches by their $\delta$ scores and retain the top 10 as the Decisive Patches ($\mathbf{P}_\text{D}=\{\mathbf{p}_1,...,\mathbf{p}_{10}\}$) for each image.
\\%我们基于分数对于不同pacth进行排序，并filter out top10的patch作为每张图中的Decisive patches.
\subsubsection{InfoDense Cut.} Here, we need to utilize the decisive patches $\mathbf{P}_\text{D}$ to guide the cutting extraction of the most info-dense partial region in each image. To achieve this, we first consider deploying ClockMix~\cite{cheng2025ed} as the cutting manner to extract regions that are dense in forgery information while maintaining semantic uniformity. Since ClockMix has already been proven to cut out segments in face forgery that evenly include identity, forgery, and background, we only need to focus on how to compute the ClockMix-segment that contains the maximum number of decisive patches in each image. Specifically, we employ a two-pointer scanning strategy. Given an angle ratio ofthe  extracted region ($n$) and two rays ($v_1$) and ($v_2$) with an angle of $\frac{360}n$, these two rays will scan all angles starting from the origin vertical one, ultimately obtaining the regional segment ($\mathbf{s}$) that contains the largest count of decisive patches as:
%With decisive attention, we then need to利用这些attention来截取每张图片中最适合回放用的infodense partial region。因此，我们想到了以clockmix作为手段，来得到forgery info 稠密，且语义均匀的区域。ClockMix已经被证明可以切出人脸伪造中，身份，伪造，和背景均匀包含的片段。因此，我们只需要考虑如何使用clockmix来切出每张图中包含decisive patch数量最大的区域。具体地说，我们采用了双指针扫描的遍历策略。给定一个region大小n（aka数据压缩比例 ）,以及对应的两条夹角固定为360/n的射线$v1,v2$, 这两条射线会从原点开始扫描所有角度，并最终记录包含的Decisive patch最多的InfoDense Segment as
\begin{equation}
    \mathbf{s}=\text{Cut}(\mathbf{x}) = \underset{\alpha \in [0,\,360-360/n)}{\arg\max}\left( \text{Count}_\text{D}(\text{Seg}(v_1,v_2,\alpha)) \right), 
\end{equation}
where $\text{Count}_\text{D}(\text{Seg})=\sum_{\mathbf{p} \in \mathbf{P}_\text{D}} \mathbf{I}\left( \mathbf{p}\text{\space within \space}\text{Seg}\right)$ represents the decisive patch count within the segment, $\mathbf{I}\left(\cdot\right)$ is an indicator that returns 1 when the input expression is true, $\alpha$ is the angle from begin to $v_1$. Based on the aforementioned algorithm, we can obtain the InfoDense Cut Segment set $\mathbf{S}=\{\mathbf{s},...\}$ corresponding to each image $\{\mathbf{x},...\} \in \mathbb{I}^{t-1}$.
%基于上述算法，我们可以得到$\mathbb{I}^{t-1}$中所有数据各自对应的InfoDense Cut Region集合$\mathbf{S}}$。
\\
\subsubsection{InfoDense Select.} Given a set of InfoDense Cut Segments, we further need to select the most information-dense one to be stored in the final replay buffer. Here, we consider both the latent-space centroid similarity~\cite{pan2023dfil} and the count of included decisive patches for the replay selection. This is because the central data of the distribution contains the most representative information of the overall dataset, while the decisive patch count represents the information that plays a critical role in forgery detection. Therefore, we combine the centroid distance and patch count with a weighted sum, which is then used for sorting. 
%一组InfoDense Cut Region,我们要进一步的选择信息丰富的块最终存入replay buffer。这里，我们同时考虑了segment的distribution centroid distance （cite DFIL）和 包含Decisive patch count来进行replay选取。这是因为分布的中心区域包含了最具数据集整体代表性的信息，而decisive patch count则代表了对于伪造判别起决定性作用的信息。具体地说，我们将中心距离和count进行了加权结合，以此作为排序的reference score。，即：
\begin{equation}
    \hat{\mathbf{S}} = \text{sort}\left(\mathbf{S}, \text{Score}(\mathbf{s}_i)\right),
\end{equation}
where the selection score of each candidate segment $\mathbf{s}_i$ is defined as
\begin{equation}
    \text{Score}(\mathbf{s}_i)
    =
    \lambda \cdot \widetilde{R}_i
    +
    (1-\lambda) \cdot \widetilde{D}_i .
\end{equation}
Here, $R_i=\cos(\mathbf{c}, f(\mathbf{x}_i))$ measures the latent-space representativeness of the source image $\mathbf{x}_i$, and $D_i=\text{Count}_{\text{D}}(\mathbf{s}_i)$ denotes the number of decisive patches covered by the segment $\mathbf{s}_i$. The centroid feature is computed as $\mathbf{c}=\mathbb{E}_{\mathbf{x}\in\mathbb{I}^{t-1}}[f(\mathbf{x})]$, where $\mathbb{E}[\cdot]$ denotes the expectation operator. Since $R_i$ and $D_i$ have different numerical ranges, we normalize both terms to $[0,1]$ over the candidate segment set before combination, yielding $\widetilde{R}_i$ and $\widetilde{D}_i$.

%最后，我们在S中选取了前m个segment存入replay buffer S^{t-1}，其中m代表设定的replay set大小。
\subsection{Replay Utilization}
\subsubsection{InfoFuse.} While conventional replay data are intact and ready for direct training, InfoDense segments are inherently fragmented and incompatible with model ingestion. Therefore, we reconstruct intact samples by fusing these fragmented replays with current-task data $\mathbb{I}^{t}$. Essentially, current-task samples serve as the base, onto which InfoDense segments are integrated to retain previous-task knowledge. Formally, for a replay segment $\mathbf{s} \in \mathbf{S}^{t-1}$, a current-task sample $\mathbf{x} \in \mathbb{I}^{t}$, and its binary mask $\mathbf{M}_{\mathbf{s}}$, the fusion is formulated as:
\begin{equation}
\mathbf{x}^{t}_{\mathrm{train}}
=
\operatorname{InfoFuse}(\mathbf{x},\mathbf{s})
=
(1-\mathbf{M}_{\mathbf{s}})\odot\mathbf{x}
+
\mathbf{M}_{\mathbf{s}}\odot\mathbf{s},
\label{eq:fuse}
\end{equation}
where $\odot$ denotes element-wise multiplication, and $\mathbf{M}_{\mathbf{s}}=1$ inside the replay region and $0$ otherwise.\\%上述过程可以被相同的应用于更之前的replay segment上。
\subsubsection{Multi-Fused Samples.}
Due to the various existence of current task data, past task data, and the combinations of their real/fake labels, the strategy behind their pairings and the labeling requires further analysis. We adopt the labeling strategy among real/fake combinations from ClockMix~\cite{cheng2025ed}, where any inclusion of ``fake'' segment is labeled as fake. This is because we want to ensure that all real/fake data in the replay buffer is effectively utilized for real/fake judgment training. Then, the combinations among previous and current tasks are newly introduced by our mixing paradigm and need to be carefully designed here. For simplicity, we use (t-1) to represent all past task data.\\
1. \textbf{(t) \& (t-1) inter-task}: We perform a full combination of the real/fake labels from (t-1) and (t) (i.e., $2\times2=4$ combinations). By mixing data with different labels for replay, we impose a constraint that helps the model learn the full information in the replay buffer and prevents class bias.\\
2. \textbf{(t) intra-task}: To ensure that the model maintains desensitization with fused images that contain junctions, we also perform random mixing of real/fake labels within the (t) task. This avoids spurious correlations where fused samples are misattributed to task (t-1), thereby preventing task bias in the model's learning process.\\
3. \textbf{Multi-Seg}: Following Eq.~\ref{eq:fuse}, past segments from task (t-1) are integrated with various current-task instances (t). This fusion maximizes the retention of historical knowledge and reinforces the supervisory signal of past data to prevent forgetting. Additionally, such multi-segment fusion boosts sample diversity and captures finer-grained subtle forgery artifacts.

All of the aforementioned types of fused samples, along with the original images from the (t)-task, form a mini-batch for training the (t)-task model ($f(\cdot, \theta^t)$).
In summary, Multi-Fused samples have the following three advantages:
1. They allow fragmented regional replay data to be effectively utilized in an intact form for replay training.
2. The fusion strategy prevents class and task biases that could arise from the introduction of fusion among different tasks and labels.
3. Fused samples contain more diverse and fine-grained information, enhancing the model's ability to learn generalizable and subtle knowledge.

%由于存在当前task数据，过往task数据，以及真假等多种组合情况，其两两搭配的合理性与label的分配策略需要进一步的分析。 为了方便叙述，我们用t-1来代表所有过往任务的数据。（1）t\&(t-1) inter-task：这里，我们将t-1的real/fake与t的real/fake进行了全量的组合（即包含2x2=4种），并沿用cite中的包含fake即是fake的labeling策略。这是因为我们希望确保replay buffer中的所有real/fake都被有效利用，而通过不同label对于不同回放数据的混合约束，可以帮助模型更好的学习replay中的全量信息，防止class bias。（2）t intra-task：为了使模型维持对于带有fusion接缝图片的脱敏，我们也同时对t task内部的真伪进行了随机的混合，进而防止模型产生错误的suprious映射：即包含fusion的图片就被认为是t-1 task，导致模型学习的信息产生task bias。（3）Multi-Seg: 过往的t-1 segment 可以被多个fusion到t task上，使得过往task的信息含量上升，提高监督训练时过往样本的影响，降低遗忘率。Moreover，multi-seg fusion可以进一步的提高数据的多样性。上述的所有类型fused 样本将与t-task的原图一起组成mini-batch，对t-task 的模型f(,\theta^t)进行训练。

%综合来说，Multi-Fused samples 有如下四个优点：1）使不规则的regional 回放数据被以规则的形式有效用于回放训练。2）通过fusion的组合策略， 我们防止了由于引入fusion而导致的class and task bias。3）单张样本包含更diverse的信息，提高了模型学习到通用泛化信息的能力、

\begin{table*}[htbp]
    \centering
        \caption{Performance comparison for face forgery detection based on Protocol 1. The two “Avg.” values denote the average detection AUC over the incremental datasets and the cross-dataset evaluation sets, respectively. The best are denoted in bold.}
    \resizebox{\textwidth}{!}{%
        \begin{tabular}{l|c|ccc|cc|cccc|c}
            \toprule
\multirow{2}{*}{\textbf{Method}} & 
\multirow{2}{*}{\textbf{Venue}} & 
\multicolumn{5}{c|}{\textbf{Incremental Dataset}} & 
\multicolumn{5}{c}{\textbf{Cross-dataset}} \\

\cmidrule(lr){3-7}\cmidrule(lr){8-12}

 &  & 
\textbf{FF++} & \textbf{DFDCP} & \textbf{CDF} & \textbf{PD $\downarrow$} & \textbf{Avg.} & \textbf{SDv21} & \textbf{UADFV} & \textbf{WDF} & \textbf{DFD} & 
\textbf{Avg.} \\
            \midrule
            L-Bound & -- & 64.51 & 61.10 & 83.42 & 11.91 & 69.67 & 43.36 & 88.31 & 51.90 & 63.87 & 61.86 \\ \midrule
 LwF~\cite{li2017learning}& TPAMI' 17 & 65.32& 71.48& 76.91& 10.86& 71.24& 50.12& 87.45& 74.38& 69.24&70.30\\
 iCaRL~\cite{rebuffi2017icarl} & CVPR' 17 & 62.75& 74.36& 85.92& 10.91& 74.34& 58.47& 86.02& 60.73& 74.88&70.03\\
 DER~\cite{yan2021dynamically} & CVPR' 21 & 64.03& 79.77& 94.68& 7.35& 79.49& 66.91& 85.10& 65.89& 80.56&74.62\\
 CoReD~\cite{kim2021cored} & MM' 21 & 66.84& 82.15& 91.37& 6.92& 80.12& 72.84& 84.92& 68.77& 85.34&77.97\\
DFIL~\cite{pan2023dfil} & MM' 23 & 77.66 & 86.48 & 95.45 & 6.48 & 86.53 & 76.53 & 84.87 & 70.01 & 87.03 & 79.61 \\
 HDP~\cite{sun2025continual} & IJCV' 24 & 73.34& 81.96& 92.90& 6.59& 82.73& 60.08& 80.82& 72.59& 80.01&73.38\\
            SUR-LID~\cite{cheng2025stacking} & CVPR' 25 & 83.21 & 80.81 & 94.85 & 6.87 & 86.29 & 72.02 & 93.43 & 66.50 & 67.50 & 74.86 \\
 GPL~\cite{gpl}& ICCV'25& 77.92& 85.43& 91.99& 5.91& 85.11& 83.05& \textbf{95.17}& 72.12& 80.09&82.61\\ \midrule
            Ours & -- & \textbf{84.11} & \textbf{89.41} & \textbf{97.75} & \textbf{4.52} & \textbf{90.42} & \textbf{97.88} & 94.71 & \textbf{79.38} & \textbf{88.13} & \textbf{90.03} \\
            \bottomrule
        \end{tabular}%
    }
    \label{tab:comparison}
\end{table*}

\begin{table*}[t]
    \centering
    \tiny
        \caption{Ablation study on compression strategies. \textbf{nSeg} denotes the average number of decisive patches per replay segment.}
    \resizebox{\textwidth}{!}{%
        \begin{tabular}{l|c|ccc|cc|cccc|c}
            \toprule
\multirow{2}{*}{\textbf{Method}} & 
\multirow{2}{*}{\textbf{nSeg}} & 
\multicolumn{5}{c|}{\textbf{Incremental Dataset}} & 
\multicolumn{5}{c}{\textbf{Cross-dataset}} \\

\cmidrule(lr){3-7}\cmidrule(lr){8-12}

 &  & 
\textbf{FF++} & \textbf{DFDCP} & \textbf{CDF} & \textbf{PD $\downarrow$} & \textbf{Avg.} & \textbf{SDv21} & \textbf{UADFV} & \textbf{WDF} & \textbf{DFD} & 
\textbf{Avg.} \\
            \midrule
            Base(DFIL) & -- & 77.66 & 86.48 & 95.45 & 6.48 & 86.53 & 43.36 & 88.31 & 51.90 & 63.87 & 61.86 \\
            JPEG (Q=50) & -- & 73.22 & 89.11 & 94.34 & 4.84 & 85.55 & 71.99 & 88.24 & 70.54 & 82.73 & 78.37 \\
            JPEG (Q=95) & -- & 79.26 & 84.03 & 91.45 & 5.16 & 84.91 & 79.29 & 94.22 & 79.28 & 91.40 & 86.04 \\
            CIM\cite{luo2023class} & -- & 83.12 & 87.18 & 94.94 & 4.97 & 88.41 & 89.04 & 94.60 & 71.82 & 82.20 & 84.41 \\
            CutMix & 3.86 & 72.34 & 76.24 & 90.38 & 6.72 & 79.65 & 57.14 & 93.52 & 75.21 & 73.98 & 74.96 \\
            CutMix+$\mathbf{P}_\text{D}$ & 7.52 & 79.83 & 83.13 & 93.94 & 6.31 & 85.63 & 62.70 & 87.29 & 68.31 & 84.46 & 75.69 \\
            ClockMix & 3.34 & 78.12 & 87.00 & 93.85 & 6.01 & 86.32 & 83.40 & 84.56 & 73.78 & 81.20 & 80.58 \\
            InfoDense (Ours) & 8.31 & 84.11 & 89.41 & 97.75 & 4.52 & 90.42 & 97.88 & 94.71 & 79.38 & 88.13 & 90.03 \\
            \bottomrule
        \end{tabular}%
    }
    \label{tab:ablation}
\end{table*}
\section{Experimental Results}

\subsection{Experimental Settings} \label{sec:exp_set}
\subsubsection{Datasets.}
To construct a comprehensive benchmark that covers diverse domains and forgery techniques, we utilize a wide range of datasets. For classical benchmarks, we adopt FaceForensics++ (FF++)~\cite{FF++}, DeepFake Detection Challenge Preview (DFDCP)~\cite{dolhansky2020deepfake}, Celeb-DF-v2 (CDF)~\cite{li2020celeb}, DeepFake Detection (DFD)~\cite{DFD}, WildDeepfake (WDF)~\cite{zi2020wilddeepfake}, and UADFV \cite{li2018ictu}. To account for emerging generative threats, we further incorporate DiT and SDv21 from DF40~\cite{df40}, and SDv21 from DiffusionFace~\cite{diffusionface}. Although two datasets share the name SDv21, they originate from different sources. In our IFFD protocol, we use the SDv21 version from DiffusionFace~\cite{diffusionface} for training and evaluation, while the SDv21 version from DF40~\cite{df40} is reserved as a hold-out set to assess generalization ability.
% 数据集。为了构建一个涵盖不同领域和伪造技术的综合基准，我们使用了广泛的数据集。
% 对于经典基准，我们使用了 FaceForensics++ (FF++)、DFDCP、Celeb-DF-v2 (CDF)、DeepFake Detection (DFD)、WildDeepfake (WDF) 和 UADFV。
% 为了应对新兴的生成式威胁，我们引入了 DF40 数据集中的 DiT 和 SDv21，以及 DiffusionFace 数据集中的 SDv21。
% 值得注意的是，虽然有两个数据集都命名为 SDv21，但我们特指在增量学习协议中使用 DiffusionFace 版本，而 DF40 版本则作为保留数据集用于泛化性评估。

\subsubsection{Incremental Protocols.} 
We design two complementary incremental evaluation protocols to systematically assess the model’s plasticity and stability under evolving forgeries:
% 增量协议。我们引入了两个不同的协议来评估模型在不断演变的伪造威胁下的可塑性和稳定性：
\begin{itemize}
    \item \textbf{Protocol 1:} 
    \textbf{Training:} The model is incrementally trained on the sequence \{FF++, DFDCP, CDF, SDv21 (DiffusionFace)\}. 
    \textbf{Testing:} After completing training on SDv21 (DiffusionFace), the resulting model is regarded as the final model. Forgetting is evaluated by testing this final model on the previously seen datasets (FF++, DFDCP, and CDF). \textit{Generalization} is assessed by evaluating its performance on SDv21 (DF40) as a held-out diffusion benchmark, as well as additional unseen datasets including UADFV, WDF, and DFD.

    % 协议 1：顺序为 {FF++, DFDCP, CDF, SDv21}。该协议混合了经典高质量数据集与最新的基于扩散模型的伪造数据，作为综合基准。
    
    \item \textbf{Protocol 2:} Sequence of \{DFDCP, SDv21 (DiffFace), DiT, CDF\} is introduced to simulate diverse domain shifts and heterogeneous generation architectures, e.g., Diffusion Transformers (DiT), thereby imposing a rigorous evaluation of the detector’s robustness and adaptability.
    % 协议 2：顺序为 {DFDCP, SDv21, DiT, CDF}。该设置引入了变化的域差距和生成架构（例如 DiT 中的扩散 Transformer），对检测器提出了严格的考验。
\end{itemize}

\subsubsection{Implementation Details.} 
Our method is implemented in PyTorch with the pre-trained CLIP ViT-L/14 as the backbone network. All input images are resized to $224 \times 224$ to satisfy the encoder requirements. The model is optimized using the Adam optimizer with an initial learning rate of $2\times10^{-4}$, a batch size of 8, and is trained for 5 epochs per task. All experiments are conducted on a single NVIDIA Tesla A100 GPU. For evaluation, frame-level Area Under the Curve (AUC) is deployed as the primary metric to measure detection performance. To further quantify catastrophic forgetting, we introduce a Performance Dropping rate (PD), defined as $PD = M_0 - M_N$, where $M_0$ represents the average AUC in the base session and $M_N$ represents the average AUC in the final session.
%实现细节。我们的方法基于 PyTorch 实现。我们采用预训练的 CLIP ViT-L/14 作为骨干网络。因此，所有输入图像均被调整大小为 224x224 以符合编码器的要求。模型使用 Adam 优化器进行优化，初始学习率为 2e-4。我们将批次大小设置为 8，每个任务训练 5 个 Epoch。所有实验均在 NVIDIA Tesla A100 和 GeForce RTX 3090 GPU 上进行。对于评估指标，我们报告帧级 AUC 作为检测准确性的主要指标。此外，我们定义了性能下降率 (PD) 来量化灾难性遗忘导致的绝对性能下降，计算公式为 $PD = M_0 - M_N$，其中 $M_0$ 是基础阶段（base session）的平均指标（AUC），$M_N$ 是最终阶段（final session）的平均指标。
\begin{table*}[t]
    \centering
    \tiny
    \caption{Ablation study on regional selection strategies. \textbf{nSeg} denotes the average decisive patches per replay segment.}
    \resizebox{\textwidth}{!}{%
        \begin{tabular}{l|c|ccc|cc|cccc|c}
            \toprule
\multirow{2}{*}{\textbf{Method}} & 
\multirow{2}{*}{\textbf{nSeg}} & 
\multicolumn{5}{c|}{\textbf{Incremental Dataset}} & 
\multicolumn{5}{c}{\textbf{Cross-dataset}} \\

\cmidrule(lr){3-7}\cmidrule(lr){8-12}

 &  & 
\textbf{FF++} & \textbf{DFDCP} & \textbf{CDF} & \textbf{PD $\downarrow$} & \textbf{Avg.} & \textbf{SDv21} & \textbf{UADFV} & \textbf{WDF} & \textbf{DFD} & 
\textbf{Avg.} \\
            \midrule
            Random      & 3.34 & 79.02 & 78.51 & 92.23 & 6.58 & 83.25 & 83.40 & 84.56 & 73.78 & 81.20 & 80.58 \\
            Center      & 6.46 & 80.26 & 84.20 & 94.93 & 5.80 & 86.46 & 84.87 & 89.32 & 72.86 & 76.58 & 80.90 \\
            Center+Hard & 6.64 & 80.46 & 83.15 & 96.11 & 5.56 & 86.57 & 79.00 & 83.74 & 70.48 & 83.80 & 79.25 \\
            Ours        & 8.31 & 84.11 & 89.41 & 97.75 & 4.52 & 90.42 & 97.88 & 94.71 & 79.38 & 88.13 & 90.03 \\
            \bottomrule
        \end{tabular}%
    }
    \label{tab:selection_strategy}
\end{table*}

\begin{table*}[t]
    \centering
    \tiny
    \caption{Ablation study on inter-task and intra-task fusion strategies. $({t-1})\&({t-1})$ and $(t-1)\&(t)$ denote isolated historical self-fusion and inter-task fusion without intra-task mixing, respectively.}
        \resizebox{\textwidth}{!}{%
        \begin{tabular}{l|ccc|cc|cccc|c}
            \toprule
\multirow{2}{*}{\textbf{Method}} & 

\multicolumn{5}{c|}{\textbf{Incremental Dataset}} & 
\multicolumn{5}{c}{\textbf{Cross-dataset}} \\

\cmidrule(lr){2-6}\cmidrule(lr){7-11}

 &  
\textbf{FF++} & \textbf{DFDCP} & \textbf{CDF} & \textbf{PD $\downarrow$} & \textbf{Avg.} & \textbf{SDv21} & \textbf{UADFV} & \textbf{WDF} & \textbf{DFD} & 
\textbf{Avg.} \\
            \midrule
            $({t-1})\&({t-1})$     & 63.75 & 61.27 & 80.11 & 12.23 & 68.37 & 63.29 & 86.65 & 59.88 & 66.63 & 69.11 \\
            $(t-1)\&(t)$     & 79.23 & 89.75 & 94.72 & 4.91  & 87.90 & 72.74 & 96.19 & 67.62 & 80.74 & 79.32 \\
            Ours  & 84.11 & 89.41 & 97.75 & 4.52  & 90.42 & 97.88 & 94.71 & 79.38 & 88.13 & 90.03 \\
            \bottomrule
        \end{tabular}%
    }
    \label{tab:fusion_strategy}
\end{table*}
\subsection{Comparison with IFFD Baselines}

% 表格 1：协议 1 上的 SOTA 方法性能对比。
% 注意：为了简洁起见，第 4 个增量任务（来自 DiffusionFace 的 SDv21）未明确显示。
% SDv21 列指的是来自 DF40 的未见数据集。
% PD 表示平均性能下降率。
% 第一个 Avg. 代表三个历史任务的平均准确率（Pre-Avg），不包括最终阶段。

We evaluate InfoDense against state-of-the-art incremental face forgery detection methods under Protocol 1 to assess continual learning stability and cross-domain generalization. As shown in Tab.~\ref{tab:comparison}, InfoDense achieves the lowest PD, indicating strong resistance to catastrophic forgetting on previously learned datasets, including FF++, DFDCP, and Celeb-DF. Unlike methods that replay complete historical images with limited diversity, InfoDense removes task-irrelevant background regions, enabling more diverse replay under the same memory budget and improving stability across incremental tasks.

Additionally, InfoDense guides the model to focus on concentrated and decisive forgery fragments rather than entire facial regions, encouraging intrinsic and transferable forgery cues instead of dataset-specific biases. Consequently, it achieves clear \textbf{\textit{cross-domain gains}} and significantly outperforms competitors such as SUR-LID on challenging unseen datasets like WDF. More detailed comparisons are provided in Appendix Tab.~\ref{tab:dmp_devfd_ours} and Fig.~\ref{fig:aaaf}.

% 我们在协议 1 上将提出的 InfoDense 与最先进的增量人脸伪造检测方法进行了评估。不同于现有方法静态存储完整历史图像并受限于回放多样性，InfoDense 主动剥离了与任务无关的冗余背景，从而在完全相同的内存预算下实现了显著更高的回放多样性。这种扩展的历史分布使我们的方法在 FaceForensics++、DFDCP 和 Celeb-DF 等先前任务上实现了最低的性能下降率和卓越的稳定性。更重要的是，通过迫使网络不断重温高度集中的决定性片段而不是完整人脸，模型捕捉到了本质的、可迁移的伪造痕迹，而不是过拟合于特定数据集的域偏差。这种范式转变为未见领域带来了卓越的泛化能力，特别是在具有挑战性的野外数据集 WDF 上大幅超越了 SUR-LID 等竞争对手。

\subsection{Ablation Study on Compression Strategies}

\subsubsection{Comparison with Global Compression Baselines.}
A comparison with global compression baselines, including JPEG at different compression ratios and CIM, is presented in Tab.~\ref{tab:ablation}. Moderate global compression enables more historical samples to be stored and brings a slight performance improvement over the uncompressed baseline. However, as the compression ratio increases, subtle high-frequency artifacts that are critical for deepfake detection are progressively destroyed, resulting in significant performance degradation. This phenomenon highlights the intrinsic limitation of global lossy compression, which reduces memory usage at the expense of indiscriminately damaging essential forgery traces. In contrast, InfoDense avoids this trade-off by adopting spatial compression rather than quality compression. Task-irrelevant background regions are removed, while the original, uncompressed pixels in information-dense areas are strictly preserved. This design maintains high-fidelity forgery cues and ensures more robust knowledge preservation under the same memory budget.

% \subsubsection{与压缩基线的比较}
% 我们首先将我们的方法与包括不同压缩比率的 JPEG 和 CIM 在内的全局压缩基线进行了评估。虽然适度的全局压缩允许存储更多历史样本，并在最初比未压缩的基线获得了性能提升，但进一步提高压缩比率不可避免地会破坏对深度伪造检测至关重要的微小高频伪影，导致严重的性能下降。这暴露了全局有损压缩的固有缺陷，即它不加区分地牺牲关键伪造痕迹的保真度来节省内存空间。形成鲜明对比的是，InfoDense 通过执行空间压缩而非质量压缩，完全规避了这一困境。通过完全丢弃与任务无关的背景，同时严格保留信息密集区域未经压缩的原始像素，我们的策略确保了核心伪造线索的高保真保留，从而实现了最稳健的知识保存。

\begin{figure}[t]
\centering
\begin{minipage}{0.28\textwidth}
  \centering
  \includegraphics[width=\linewidth]{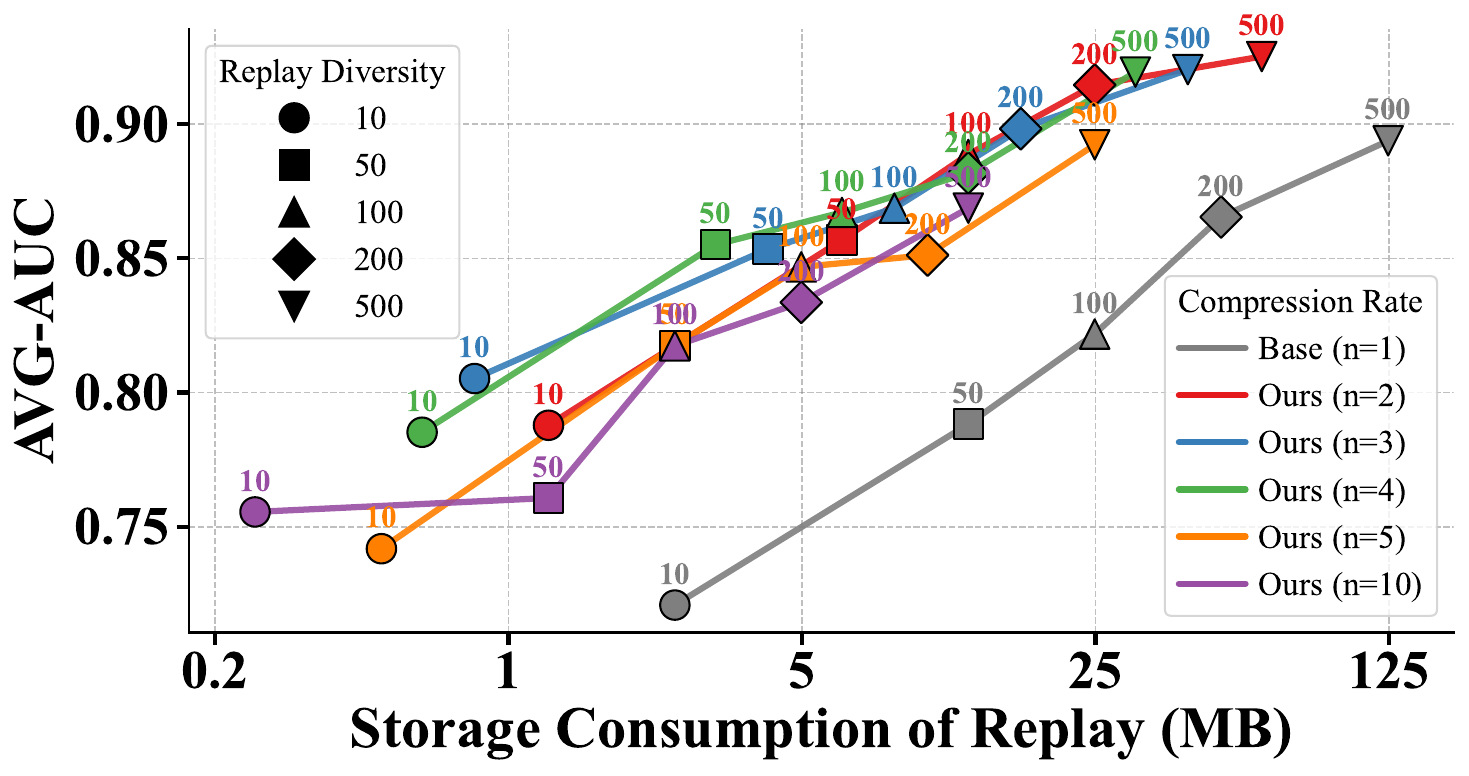}
\end{minipage}
\hfill
\begin{minipage}{0.17\textwidth}
  \centering
  \includegraphics[width=\linewidth]{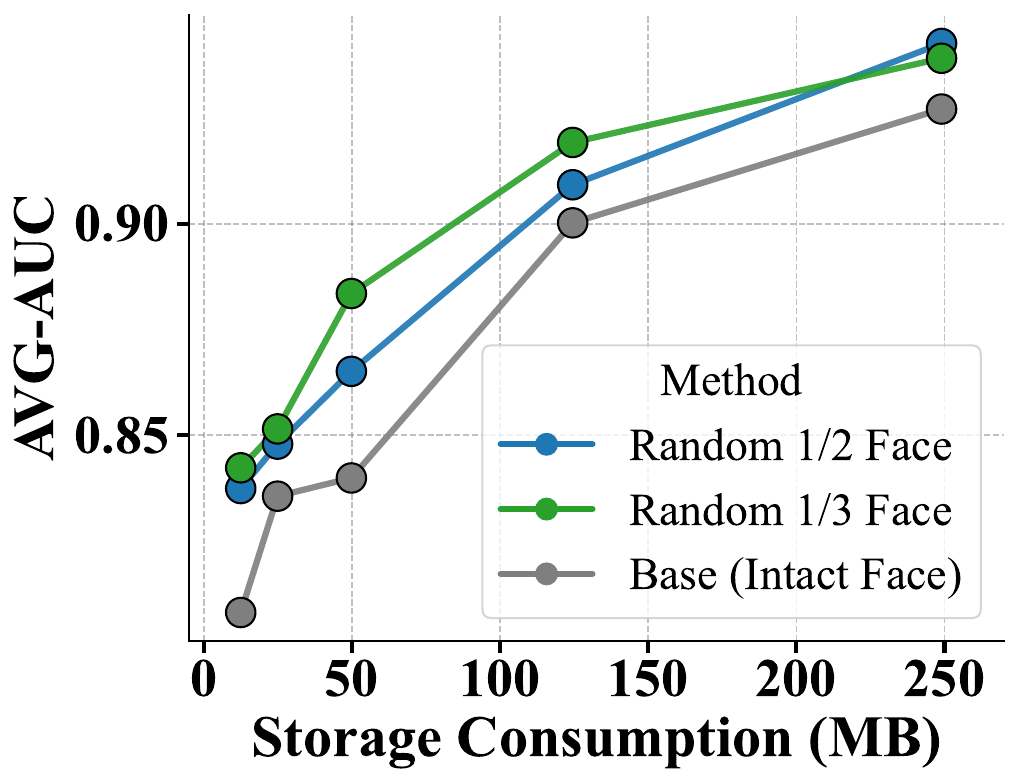}
\end{minipage}
\caption{Storage-performance Pareto frontier under varying replay budgets. (Left) Results evaluated on Protocol 1. (Right) Results evaluated on Protocol 2.}

% 图 4：不同回放预算下的存储-性能 Pareto 前沿。（左图）在 Protocol 1 上的评估结果。（右图）在 Protocol 2 上的评估结果。
\label{fig:storage}
\end{figure}

\subsubsection{Impact of Regional Shapes and Patch Selection.}

We analyze region shapes and patch selection strategies in Tab.~\ref{tab:ablation}, where Cut/Clock denote rectangular/fan-shaped regions, Mix denotes random sampling, and $P_D$ or InfoDense uses decisive patch density for selection.\\
\textbf{1. Rectangular vs.\ Fan-shaped.}
Under random sampling, fan-shaped ClockMix consistently outperforms rectangular CutMix. This is because facial structures and forgery artifacts often follow a center-symmetric distribution, making sectors radiating from the facial center more likely to preserve key structural and manipulation cues, while random rectangles may miss the facial center.\\
\textbf{2. Random vs.\ Density-Aware Patch Selection.}
Density-aware selection with decisive patches $P_D$ clearly improves information concentration. Random sampling covers few decisive regions, as reflected by the low nSeg values of CutMix and ClockMix. In contrast, CutMix+$P_D$ and InfoDense substantially increase decisive-region coverage. By combining center-symmetric fan-shaped regions with forgery-cue density, InfoDense achieves a better trade-off between spatial compression and discriminative performance.

% \subsubsection{区域形状与提取引导机制的影响。}
% 我们进一步研究了区域形状（包括矩形和扇形）以及提取引导策略（包括随机裁剪和我们提出的密度感知提取）的影响。在表 2 中，Cut 和 Clock 分别表示基础的矩形和扇形区域。Mix 代表随机提取，而引入 P_D 以及我们提出的 InfoDense 则是利用决定性 Patch 的密度来主动引导提取过程。
    
    % 矩形 vs. 扇形：当比较随机提取基线时，扇形的 ClockMix 天然优于矩形的 CutMix。由于面部特征和伪造痕迹通常表现出中心对称的分布，从面部中心辐射出的扇形区域本质上能捕捉到更多关键的结构信息。相反，随机的矩形裁剪很容易完全错过关键的面部中心，导致较差的知识保留。

    % 随机 vs. 密度感知提取：利用决定性 Patch P_D 进行提取引导的必要性非常明显。仅依赖随机裁剪会导致决定性 Patch 的包含量极低，这从 CutMix 和 ClockMix 较低的 nSeg 指标中可以看出。通过引入 P_D 的密度来引导提取，CutMix+P_D 和我们的 InfoDense 在信息密度上都实现了大幅提升。最终，我们的 InfoDense 策略完美地将中心对称的扇形提取与伪造线索的内在空间密度对齐，在空间压缩和判别性能之间实现了最佳平衡。

% 表格 2：压缩策略消融实验。
% 注意：为了简洁起见，第 4 个增量任务未明确显示。
% SDv21 列指的是来自 DF40 的未见数据集。
% PD 表示平均性能下降率。
% 第一个 Avg. 代表三个历史任务的平均准确率（Pre-Avg）。
% Cvg. (覆盖率) 表示回放区域中包含的前 10 个关键 Patch (top-10 decisive patches) 的平均数量，该指标是在 FF++ 数据集上计算得出的。
\begin{figure}[t]
\centering
\begin{minipage}{0.23\textwidth}
  \centering
  \includegraphics[width=\linewidth]{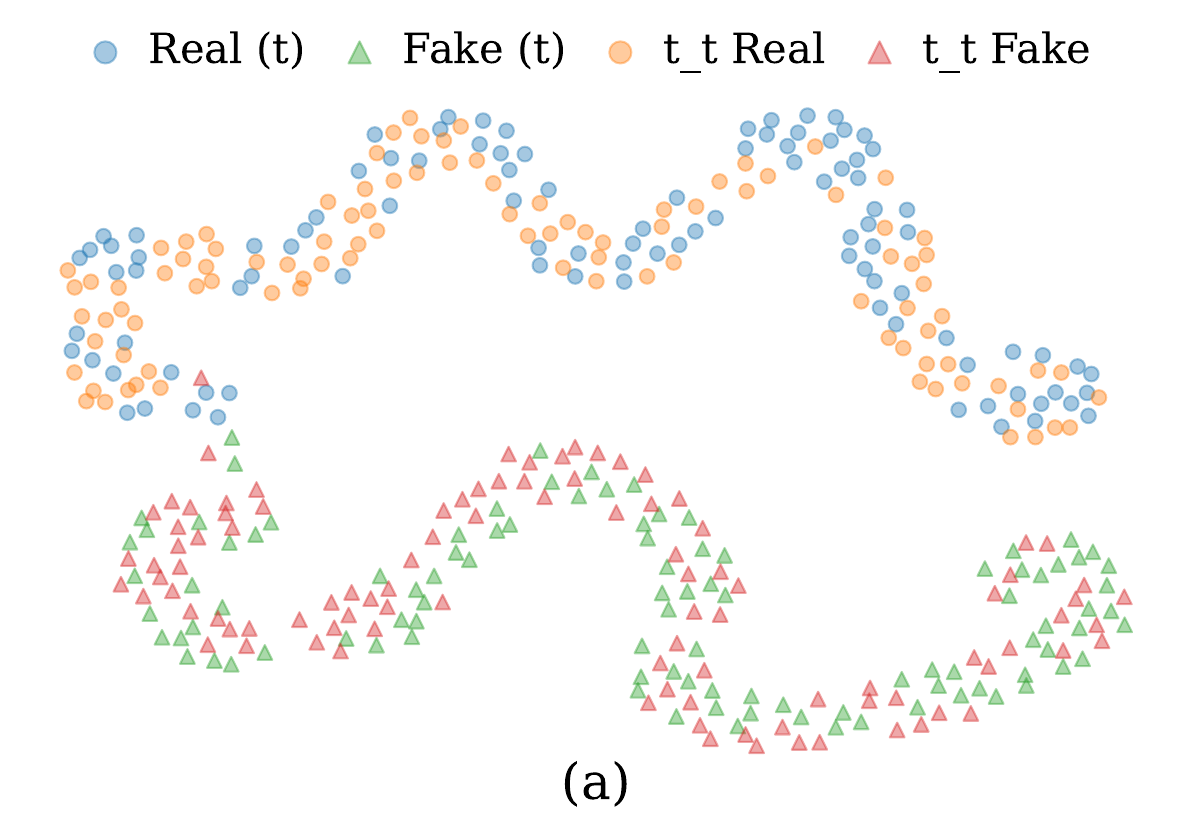}
\end{minipage}
\hfill
\begin{minipage}{0.23\textwidth}
  \centering
  \includegraphics[width=\linewidth]{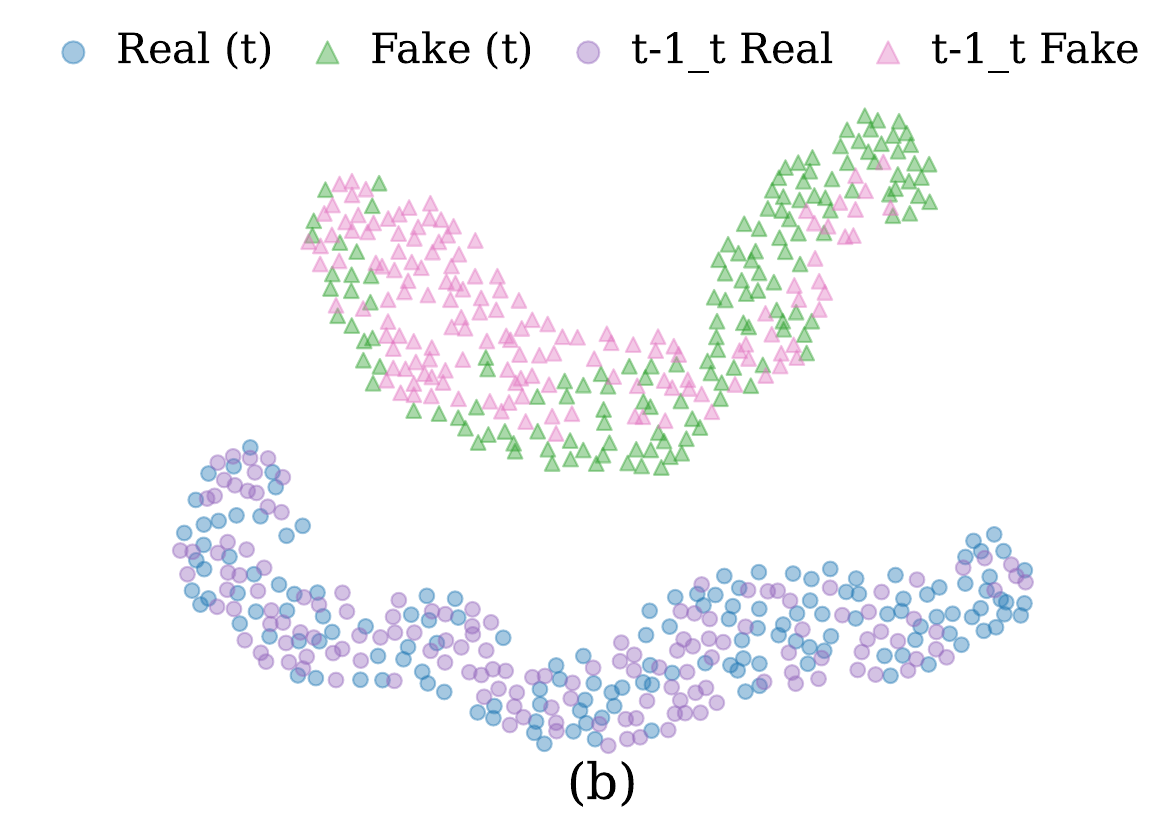}
\end{minipage}
\caption{UMAP visualization of the latent space. (a) Original Task $t$ samples \textit{vs.} intra-task self-fused samples. (b) Original Task $t$ samples \textit{vs.} inter-task replay samples.}
\label{fig:vis}
\vspace{-0.6cm}
\end{figure}
\begin{figure*}[t]
    \centering
    \includegraphics[width=1\linewidth]{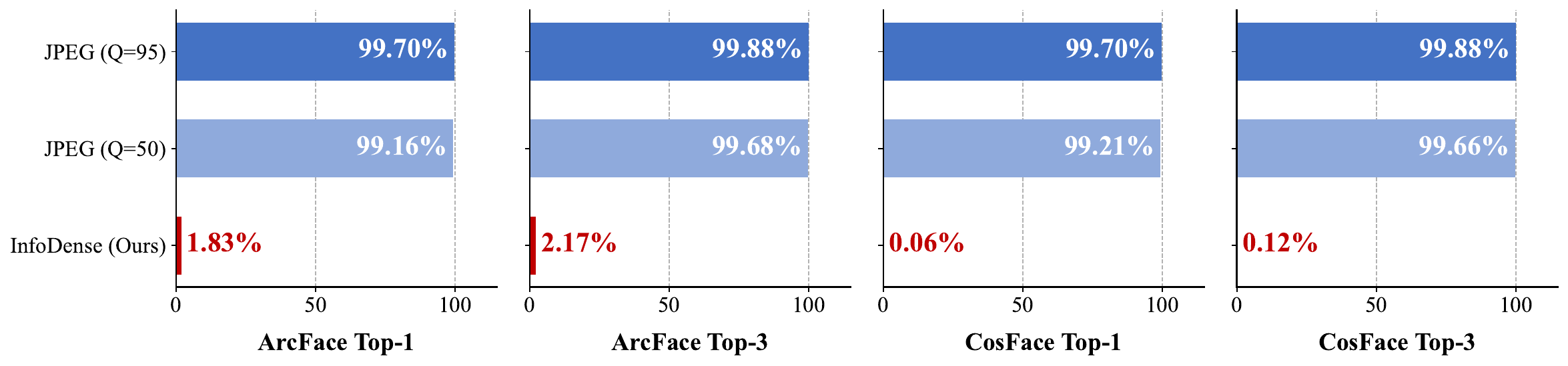}
    \caption{Quantitative analysis of identity privacy protection. Top-1 and Top-3 recall rates denote the probability of retrieving original identities from replay samples via face recognition (\textit{e.g.}, ArcFace, CosFace). Lower rates imply less identity leakage.}
    % 身份隐私的定量分析。Top-1 和 Top-3 召回率表示使用最先进的人脸识别模型从存储的回放样本中成功检索出原始身份的概率。
    \label{fig:privacy}
   
\end{figure*}
\subsection{Ablation on Selection and Fusion Strategies}

\subsubsection{Effectiveness of Selection Strategy.}

Tab.~\ref{tab:selection_strategy} compares InfoDense with Random, Center, and Center+Hard sampling. For Center, the decisive patch count is set to zero, and selection relies only on latent-space centroid distance to preserve distribution representativeness. Although central samples and decision-boundary hard samples slightly outperform random selection, they cannot ensure that sampled fragments contain forgery artifacts, yielding insufficient task-relevant cues.
In contrast, InfoDense dynamically combines centroid distance with decisive patch count, filtering redundant backgrounds and enriching the replay buffer with artifact-rich fragments. This dual-metric strategy mitigates catastrophic forgetting and provides higher-quality replay data, leading to stronger generalization across unseen domains.

% \subsubsection{选择策略的有效性}
% 为了评估 InfoDense 选择策略的有效性，我们将它与 Random、Center 和 Center+Hard 基线采样方法进行了比较。Center 策略作为一个关键的消融实验，实际上是将决定性 Patch 计数的权重完全设为零，仅依赖潜在空间的质心距离来保证分布的代表性。虽然优先选择中心或靠近决策边界的困难样本比随机选择有微小的提升，但这些策略在构建回放缓冲区时，盲目地忽略了裁剪片段中是否物理上存在伪造伪影。因此，它们无法捕获足够的任务相关线索。通过将质心距离与决定性 Patch 的显式计数动态结合，我们的 InfoDense 策略刻意过滤掉了冗余背景，并确保回放缓冲区内密集充斥着核心的伪造证据。这种双指标选择不仅大幅减少了灾难性遗忘，还提供了高质量、富含伪影的演练数据，从而解锁了在各种未见领域上的强大泛化能力。

\subsubsection{Effectiveness of Fusion Strategy.}

To validate InfoFuse, we compare it with two simplified variants: isolated historical self-fusion ($({t-1})\&({t-1})$) and inter-task-only fusion ($(t-1)\&(t)$). As shown in Tab.~\ref{tab:fusion_strategy}, isolated self-fusion causes severe catastrophic forgetting, as reconstructing replay samples only from historical data introduces a large domain gap from pristine current-task images and amplifies task bias. Inter-task-only fusion, which pastes historical segments onto current images, partially reduces this gap but remains sub-optimal. Without intra-task mixing on current data, the model may exploit artificial fusion boundaries as shortcuts for task identity instead of learning intrinsic forgery patterns. In contrast, InfoFuse jointly performs inter-task and intra-task mixing, exposing the model to fusion boundaries across data distributions. This reduces boundary sensitivity, encourages attention to genuine forgery traces, and improves both knowledge retention and generalization.

% \subsubsection{融合策略的有效性}
% 为了验证我们的 InfoFuse 策略，我们将它与孤立的历史自我融合（$t-1_{t-1}$）和仅跨任务融合（$t-1_t$）进行了比较。孤立的自我融合会导致严重的灾难性遗忘，因为它在重建的回放样本和原始当前数据之间产生了明显的领域鸿沟，严重诱发了任务偏差。相反，直接将历史片段粘贴到当前图像上弥合了这一鸿沟，但泛化能力依然次优。在没有对当前数据应用任务内混合的情况下，网络极易利用人为的融合边界作为区分任务的虚假捷径。我们全面的 InfoFuse 通过同时执行跨任务和任务内混合解决了这一缺陷。通过让网络在所有数据分布中暴露于融合边界，InfoFuse 迫使模型对拼接伪影脱敏并纯粹关注内在伪造线索，从而确保了稳健的知识保留和强大的泛化能力。
% \subsection{存储效率与冗余性分析}
% 我们从存储-性能权衡和经验冗余极限两个角度，进行了广泛的实验来分析我们方法的存储效率，并验证人脸伪造数据中的内在冗余假设。

\subsection{Storage Efficiency and Redundancy}

Here, we extensively analyze InfoDense storage efficiency and validate the intrinsic redundancy hypothesis in IFFD from two aspects: storage-performance trade-off and empirical redundancy limits.

\subsubsection{Performance Evaluation on Storage Efficiency.}
In practical deployment, maximizing detection accuracy under limited storage is a key Pareto optimization problem. We compare InfoDense with the full-image replay baseline DFIL under different compression ratios. As shown in Fig.~\ref{fig:storage}, InfoDense consistently achieves a superior Pareto frontier. Under the same memory budget, storing more compressed, information-dense fragments significantly outperforms storing fewer complete images. For instance, fragments at one quarter of the original size achieve higher accuracy than storing the same number of intact images while using much less memory. These results show that density-aware selection removes redundant background content and preserves richer forgery cues under strict storage constraints.

\subsubsection{Verification of Facial Redundancy.}
To empirically validate the redundancy of face forgery data, we conduct a supplementary experiment with random fan-shaped cropping on a naive CNN backbone (Fig.~\ref{fig:storage} Right). By discarding strategic selection and randomly extracting regional segments, we assess the structural resilience of forgery cues. Surprisingly, random extraction consistently outperforms full-image replay, confirming that forgery traces are highly redundant and center-symmetric. This supports our motivation that storing entire faces limits replay diversity, while even randomly sampled partial regions can retain sufficient discriminative information for robust incremental replay.

% \subsubsection{人脸冗余性的验证}
% 为了从经验上验证人脸伪造数据的内在冗余性，我们在标准 CNN 骨干网络上利用随机扇形裁剪策略进行了一项补充实验。与密度感知的 InfoDense 不同，该设置完全放弃了策略性选择过程，而是随机提取局部扇形区域以评估伪造线索的结构韧性。令人惊讶的是，性能曲线显示这种纯随机提取始终优于全图基线。这种反直觉的优越性证实了我们的核心动机，即伪造痕迹是高度冗余且中心对称的。它提供了强有力的经验证据，表明存档整张人脸从根本上限制了回放缓冲区的潜在多样性，而即使是随机采样的局部区域，本质上也封装了足够的判别信息来驱动更稳健的增量演练。

%图像退化的鲁棒性。为了验证所提出的区域决定性回放机制没有对高质量的干净数据产生过拟合，我们评估了模型在面对现实世界中常见图像退化时的稳定性。在推理阶段，我们独立地施加了四种类型的扰动，包括下采样、饱和度改变、块状丢弃和 JPEG 压缩。每种退化均配置了五个递进的强度等级，以全面评估性能的衰减趋势。我们评估了四个标准深伪数据集的平均检测性能。如图~\ref{fig:robustness} 所示，与基线模型相比，我们的方法在面对不同程度的破坏时始终表现出更优越的抗干扰能力。这证明了通过选择性地回放信息密集的区域，模型避免了依赖脆弱的伪影，并成功学习到了能够抵御严重图像破坏的鲁棒且本质的伪造痕迹。

\subsection{Visualization Analysis on InfoDense Fuse}

To show that segment fusion preserves deepfake discriminability without seaming-induced confusion, we visualize the latent space using UMAP \cite{umap}. Fig.~\ref{fig:vis}(a) shows original Task $t$ samples and intra-task self-fused samples. The fused features align well with the original real and fake clusters, indicating that fusion preserves intrinsic class distributions. Fig.~\ref{fig:vis}(b) compares original Task $t$ samples with inter-task replay samples, where Task $t-1$ segments are fused onto Task $t$. The replay features largely overlap with the corresponding real and fake domains while maintaining clear class separability. This confirms that InfoDense Fuse constructs valid replay data that preserves core forgery cues and supports historical knowledge accumulation without distorting the feature space.

%潜在空间可视化。为了直观地证明我们的片段融合策略在不引入边界混淆的情况下保持了伪造特征的判别力，我们使用 UMAP 对潜在特征空间进行了可视化。图~\ref{fig:vis}(a) 展示了原始任务 t 样本以及任务内自融合样本的特征分布。融合后的特征与其对应的原始真伪聚类无缝对齐，证明拼接操作保留了内在的类别分布。图~\ref{fig:vis}(b) 比较了原始任务 t 样本与跨任务回放样本（任务 t-1 片段融合到任务 t 上）。回放样本与其各自的真伪域高度重合，保持了严格的类别可分性。这证实了我们的方法成功构建了保留核心伪造线索的有效回放数据，从而在不扭曲特征空间的情况下实现了有效的历史知识积累。

%图~\ref{fig:vis} 潜在空间的 UMAP 可视化。(a) 原始任务 t 样本与任务内自融合样本的特征分布。(b) 原始任务 t 样本与跨任务回放样本的特征分布。

\subsection{Privacy Quantitative Analysis}
A key concern in replay-based incremental learning is privacy leakage from stored historical data. To evaluate privacy preservation, we measure the identity retrieval recall of stored samples against original faces using ArcFace \cite{deng2019arcface} and CosFace \cite{wang2018cosface}. As shown in Fig.~\ref{fig:privacy}, global compression methods still retain near-perfect retrieval recall even under high compression, as they preserve the holistic facial topology required for identity recognition. In contrast, InfoDense disrupts this structure by retaining only localized forgery-decisive fragments, preventing the recognition of coherent identity embeddings and reducing retrieval recall to near zero. This shows that InfoDense improves storage efficiency while serving as an effective de-identification mechanism for privacy-sensitive deployment. Further Privacy analysis can be found in Appendix~\ref{sec:supp-privacy}.

% 基于回放的增量学习中的一个关键问题是与存储历史用户数据相关的潜在隐私泄露风险。为了定量评估我们方法的隐私保护能力，我们使用包括 ArcFace 和 CosFace 在内的最先进人脸识别模型，测量了存储样本与原始人脸的身份检索召回率。如表~\ref{tab:privacy} 所示，传统的全局压缩方法严重损害了用户隐私，即使在较高的压缩率下也保持着近乎完美的检索召回率。这是因为全局压缩均匀地保留了对身份识别至关重要的整体面部拓扑结构。形成鲜明对比的是，InfoDense 通过有选择地隔离高度局部的伪造决定性片段，彻底破坏了这种整体结构。因此，标准的识别网络无法提取连贯的身份嵌入，导致检索召回率接近于零。这有力地证明了 InfoDense 不仅优化了存储效率，而且本质上充当了一种稳健的脱敏机制，使其非常适合对隐私敏感的实际部署。

\section{Conclusion}
In this paper, we introduce InfoDense, a Density-Aware Regional Decisive replay strategy for memory-efficient Incremental Face Forgery Detection. By actively leveraging facial image redundancy, InfoDense extracts compact, artifact-dense regions, drastically reducing overall storage costs and boosting replay sample diversity within the same memory budget. Furthermore, extracting forgery-decisive cues across continuous tasks compels the detection network to learn transferable manipulation characteristics rather than overfitting to task-specific artifacts. Extensive evaluations on multiple challenging incremental benchmarks fully validate the effectiveness of our InfoDense in maximizing knowledge retention and cross-domain generalizability.


\bibliography{AAAI2027/AnonymousSubmission2027}

\begin{thebibliography}{52}
\providecommand{\natexlab}[1]{#1}

\bibitem[{Bousselham et~al.(2025)Bousselham, Boggust, Chaybouti, Strobelt, and Kuehne}]{bousselham2025legrad}
Bousselham, W.; Boggust, A.; Chaybouti, S.; Strobelt, H.; and Kuehne, H. 2025.
\newblock Legrad: An explainability method for vision transformers via feature formation sensitivity.
\newblock In \emph{Proceedings of the IEEE/CVF International Conference on Computer Vision}, 20336--20345.

\bibitem[{Buslaev et~al.(2020)Buslaev, Iglovikov, Khvedchenya, Parinov, Druzhinin, and Kalinin}]{buslaev2020albumentations}
Buslaev, A.; Iglovikov, V.~I.; Khvedchenya, E.; Parinov, A.; Druzhinin, M.; and Kalinin, A.~A. 2020.
\newblock Albumentations: fast and flexible image augmentations.
\newblock \emph{Information}, 11(2): 125.

\bibitem[{Cao et~al.(2022)Cao, Ma, Yao, Chen, Ding, and Yang}]{cao2022end}
Cao, J.; Ma, C.; Yao, T.; Chen, S.; Ding, S.; and Yang, X. 2022.
\newblock End-to-End Reconstruction-Classification Learning for Face Forgery Detection.
\newblock In \emph{IEEE Conference on Computer Vision and Pattern Recognition}, 4113--4122.

\bibitem[{Chattopadhay et~al.(2018)Chattopadhay, Sarkar, Howlader, and Balasubramanian}]{chattopadhay2018gradpp}
Chattopadhay, A.; Sarkar, A.; Howlader, P.; and Balasubramanian, V.~N. 2018.
\newblock Grad-cam++: Generalized gradient-based visual explanations for deep convolutional networks.
\newblock In \emph{2018 IEEE winter conference on applications of computer vision (WACV)}, 839--847. IEEE.

\bibitem[{Chefer, Gur, and Wolf(2021)}]{chefer2021transformergrad}
Chefer, H.; Gur, S.; and Wolf, L. 2021.
\newblock Transformer interpretability beyond attention visualization.
\newblock In \emph{Proceedings of the IEEE/CVF conference on computer vision and pattern recognition}, 782--791.

\bibitem[{Chen et~al.(2022)Chen, Zhang, Song, Liu, and Wang}]{chen2022self}
Chen, L.; Zhang, Y.; Song, Y.; Liu, L.; and Wang, J. 2022.
\newblock Self-supervised learning of adversarial example: Towards good generalizations for deepfake detection.
\newblock In \emph{IEEE Conference on Computer Vision and Pattern Recognition}, 18710--18719.

\bibitem[{Chen et~al.(2021)Chen, Yao, Chen, Ding, Li, and Ji}]{chen2021local}
Chen, S.; Yao, T.; Chen, Y.; Ding, S.; Li, J.; and Ji, R. 2021.
\newblock Local relation learning for face forgery detection.
\newblock In \emph{AAAI Conference on Artificial Intelligence}, volume~35, 1081--1088.

\bibitem[{Chen et~al.(2024)Chen, Sun, Zhou, Lin, Sun, Cao, and Ji}]{diffusionface}
Chen, Z.; Sun, K.; Zhou, Z.; Lin, X.; Sun, X.; Cao, L.; and Ji, R. 2024.
\newblock DiffusionFace: Towards a Comprehensive Dataset for Diffusion-Based Face Forgery Analysis.
\newblock \emph{arXiv preprint arXiv:2403.18471}.

\bibitem[{Cheng et~al.(2025{\natexlab{a}})Cheng, Yan, Zhang, Hao, Ai, Zou, Li, and Wang}]{cheng2025stacking}
Cheng, J.; Yan, Z.; Zhang, Y.; Hao, L.; Ai, J.; Zou, Q.; Li, C.; and Wang, Z. 2025{\natexlab{a}}.
\newblock Stacking brick by brick: Aligned feature isolation for incremental face forgery detection.
\newblock In \emph{IEEE Conference on Computer Vision and Pattern Recognition}, 13927--13936.

\bibitem[{Cheng et~al.(2024)Cheng, Yan, Zhang, Luo, Wang, and Li}]{cheng2024can}
Cheng, J.; Yan, Z.; Zhang, Y.; Luo, Y.; Wang, Z.; and Li, C. 2024.
\newblock Can we leave deepfake data behind in training deepfake detector?
\newblock \emph{Advances in Neural Information Processing Systems}, 37: 21979--21998.

\bibitem[{Cheng et~al.(2025{\natexlab{b}})Cheng, Zhang, Zou, Yan, Liang, Wang, and Li}]{cheng2025ed}
Cheng, J.; Zhang, Y.; Zou, Q.; Yan, Z.; Liang, C.; Wang, Z.; and Li, C. 2025{\natexlab{b}}.
\newblock Ed ˆ4: Explicit data-level debiasing for deepfake detection.
\newblock \emph{IEEE Transactions on Image Processing}.

\bibitem[{Cui et~al.(2025{\natexlab{a}})Cui, Li, Luo, Zhou, and Dong}]{ForAda}
Cui, X.; Li, Y.; Luo, A.; Zhou, J.; and Dong, J. 2025{\natexlab{a}}.
\newblock Forensics Adapter: Adapting CLIP for Generalizable Face Forgery Detection.
\newblock In \emph{IEEE/CVF Conference on Computer Vision and Pattern Recognition}, 19207--19217.

\bibitem[{Cui et~al.(2025{\natexlab{b}})Cui, Li, Luo, Zhou, and Dong}]{cui2025forensics}
Cui, X.; Li, Y.; Luo, A.; Zhou, J.; and Dong, J. 2025{\natexlab{b}}.
\newblock Forensics adapter: Adapting clip for generalizable face forgery detection.
\newblock In \emph{IEEE Conference on Computer Vision and Pattern Recognition Conference}, 19207--19217.

\bibitem[{Deng et~al.(2019)Deng, Guo, Xue, and Zafeiriou}]{deng2019arcface}
Deng, J.; Guo, J.; Xue, N.; and Zafeiriou, S. 2019.
\newblock Arcface: Additive angular margin loss for deep face recognition.
\newblock In \emph{Proceedings of the IEEE Conference on Computer Vision and Pattern Recognition}, 4690--4699.

\bibitem[{Dolhansky et~al.(2020)Dolhansky, Bitton, Pflaum, Lu, Howes, Wang, and Ferrer}]{dolhansky2020deepfake}
Dolhansky, B.; Bitton, J.; Pflaum, B.; Lu, J.; Howes, R.; Wang, M.; and Ferrer, C.~C. 2020.
\newblock The deepfake detection challenge (dfdc) dataset.
\newblock \emph{arXiv preprint arXiv:2006.07397}.

\bibitem[{Dong et~al.(2023)Dong, Wang, Ji, Liang, Fan, and Ge}]{dong2023implicit}
Dong, S.; Wang, J.; Ji, R.; Liang, J.; Fan, H.; and Ge, Z. 2023.
\newblock Implicit identity leakage: The stumbling block to improving deepfake detection generalization.
\newblock In \emph{Proceedings of the IEEE Conference on Computer Vision and Pattern Recognition}, 3994--4004.

\bibitem[{Guo et~al.(2020)Guo, Zhu, Zhao, Cao, Lei, and Li}]{guo2020learning}
Guo, J.; Zhu, X.; Zhao, C.; Cao, D.; Lei, Z.; and Li, S.~Z. 2020.
\newblock Learning meta face recognition in unseen domains.
\newblock In \emph{IEEE Conference on Computer Vision and Pattern Recognition}, 6163--6172.

\bibitem[{Huang et~al.(2023)Huang, Wang, Yang, Ai, Zou, Wang, and Ye}]{huang2023implicit}
Huang, B.; Wang, Z.; Yang, J.; Ai, J.; Zou, Q.; Wang, Q.; and Ye, D. 2023.
\newblock Implicit Identity Driven Deepfake Face Swapping Detection.
\newblock In \emph{IEEE Conference on Computer Vision and Pattern Recognition}, 4490--4499.

\bibitem[{Kashiani, Talemi, and Afghah(2025)}]{kashiani2025freqdebias}
Kashiani, H.; Talemi, N.~A.; and Afghah, F. 2025.
\newblock Freqdebias: Towards generalizable deepfake detection via consistency-driven frequency debiasing.
\newblock In \emph{IEEE Conference on Computer Vision and Pattern Recognition}, 8775--8785. IEEE.

\bibitem[{Kim, Tariq, and Woo(2021)}]{kim2021cored}
Kim, M.; Tariq, S.; and Woo, S.~S. 2021.
\newblock Cored: Generalizing fake media detection with continual representation using distillation.
\newblock In \emph{ACM International Conference on Multimedia}, 337--346.

\bibitem[{Li et~al.(2020{\natexlab{a}})Li, Bao, Zhang, Yang, Chen, Wen, and Guo}]{li2020face}
Li, L.; Bao, J.; Zhang, T.; Yang, H.; Chen, D.; Wen, F.; and Guo, B. 2020{\natexlab{a}}.
\newblock Face x-ray for more general face forgery detection.
\newblock In \emph{IEEE Conference on Computer Vision and Pattern Recognition}, 5001--5010.

\bibitem[{Li, Chang, and Lyu(2018)}]{li2018ictu}
Li, Y.; Chang, M.-C.; and Lyu, S. 2018.
\newblock In ictu oculi: Exposing ai created fake videos by detecting eye blinking.
\newblock In \emph{IEEE International Workshop on Information Forensics and Security}, 1--7.

\bibitem[{Li et~al.(2020{\natexlab{b}})Li, Yang, Sun, Qi, and Lyu}]{li2020celeb}
Li, Y.; Yang, X.; Sun, P.; Qi, H.; and Lyu, S. 2020{\natexlab{b}}.
\newblock Celeb-df: A large-scale challenging dataset for deepfake forensics.
\newblock In \emph{IEEE Conference on Computer Vision and Pattern Recognition}, 3207--3216.

\bibitem[{Li and Hoiem(2017)}]{li2017learning}
Li, Z.; and Hoiem, D. 2017.
\newblock Learning without forgetting.
\newblock \emph{IEEE Transactions on Pattern Analysis and Machine Intelligence}, 40(12): 2935--2947.

\bibitem[{Liang, Shi, and Deng(2022)}]{liang2022exploring}
Liang, J.; Shi, H.; and Deng, W. 2022.
\newblock Exploring disentangled content information for face forgery detection.
\newblock In \emph{European Conference on Computer Vision}, 128--145. Springer.

\bibitem[{Luo et~al.(2023)Luo, Liu, Schiele, and Sun}]{luo2023class}
Luo, Z.; Liu, Y.; Schiele, B.; and Sun, Q. 2023.
\newblock Class-incremental exemplar compression for class-incremental learning.
\newblock In \emph{IEEE Conference on Computer Vision and Pattern Recognition}, 11371--11380.

\bibitem[{McInnes, Healy, and Melville(2018)}]{umap}
McInnes, L.; Healy, J.; and Melville, J. 2018.
\newblock Umap: Uniform manifold approximation and projection for dimension reduction.
\newblock \emph{arXiv preprint arXiv:1802.03426}.

\bibitem[{Nick~Dufour and Andrew~Gully(2019)}]{DFD}
Nick~Dufour, G.~R.; and Andrew~Gully, J. 2019.
\newblock Deep Fake Detection Dataset.
\newblock \url{https://ai.googleblog.com/2019/09/contributing-data-to-deepfake-detection.html}.

\bibitem[{Pan et~al.(2023)Pan, Yin, Wei, Lin, Ba, Liu, Wang, Cavallaro, and Ren}]{pan2023dfil}
Pan, K.; Yin, Y.; Wei, Y.; Lin, F.; Ba, Z.; Liu, Z.; Wang, Z.; Cavallaro, L.; and Ren, K. 2023.
\newblock Dfil: Deepfake incremental learning by exploiting domain-invariant forgery clues.
\newblock In \emph{ACM International Conference on Multimedia}, 8035--8046.

\bibitem[{Qian et~al.(2020)Qian, Yin, Sheng, Chen, and Shao}]{qian2020thinking}
Qian, Y.; Yin, G.; Sheng, L.; Chen, Z.; and Shao, J. 2020.
\newblock Thinking in frequency: Face forgery detection by mining frequency-aware clues.
\newblock In \emph{European Conference on Computer Vision}, 86--103. Springer.

\bibitem[{Radford et~al.(2021)Radford, Kim, Hallacy, Ramesh, Goh, Agarwal, Sastry, Askell, Mishkin, Clark et~al.}]{clip}
Radford, A.; Kim, J.~W.; Hallacy, C.; Ramesh, A.; Goh, G.; Agarwal, S.; Sastry, G.; Askell, A.; Mishkin, P.; Clark, J.; et~al. 2021.
\newblock Learning transferable visual models from natural language supervision.
\newblock In \emph{International conference on machine learning}, 8748--8763. PMLR.

\bibitem[{Rebuffi et~al.(2017)Rebuffi, Kolesnikov, Sperl, and Lampert}]{rebuffi2017icarl}
Rebuffi, S.-A.; Kolesnikov, A.; Sperl, G.; and Lampert, C.~H. 2017.
\newblock icarl: Incremental classifier and representation learning.
\newblock In \emph{IEEE Conference on Computer Vision and Pattern Recognition}, 2001--2010.

\bibitem[{Rossler et~al.(2019)Rossler, Cozzolino, Verdoliva, Riess, Thies, and Nie{\ss}ner}]{FF++}
Rossler, A.; Cozzolino, D.; Verdoliva, L.; Riess, C.; Thies, J.; and Nie{\ss}ner, M. 2019.
\newblock Faceforensics++: Learning to detect manipulated facial images.
\newblock In \emph{IEEE International Conference on Computer Vision}, 1--11.

\bibitem[{Selvaraju et~al.(2017)Selvaraju, Cogswell, Das, Vedantam, Parikh, and Batra}]{selvaraju2017gradcam}
Selvaraju, R.~R.; Cogswell, M.; Das, A.; Vedantam, R.; Parikh, D.; and Batra, D. 2017.
\newblock Grad-cam: Visual explanations from deep networks via gradient-based localization.
\newblock In \emph{Proceedings of the IEEE international conference on computer vision}, 618--626.

\bibitem[{Shiohara and Yamasaki(2022)}]{shiohara2022detecting}
Shiohara, K.; and Yamasaki, T. 2022.
\newblock Detecting deepfakes with self-blended images.
\newblock In \emph{IEEE Conference on Computer Vision and Pattern Recognition}, 18720--18729.

\bibitem[{Sun et~al.(2025)Sun, Chen, Yao, Sun, Ding, and Ji}]{sun2025continual}
Sun, K.; Chen, S.; Yao, T.; Sun, X.; Ding, S.; and Ji, R. 2025.
\newblock Continual face forgery detection via historical distribution preserving.
\newblock \emph{International Journal of Computer Vision}, 133(3): 1067--1084.

\bibitem[{Sun et~al.(2022)Sun, Yao, Chen, Ding, Li, and Ji}]{sun2022dual}
Sun, K.; Yao, T.; Chen, S.; Ding, S.; Li, J.; and Ji, R. 2022.
\newblock Dual contrastive learning for general face forgery detection.
\newblock In \emph{AAAI Conference on Artificial Intelligence}, volume~36, 2316--2324.

\bibitem[{Tian et~al.(2024)Tian, Yu, Wang, Chen, Xiao, Han, and Chai}]{tian2024dynamic}
Tian, J.; Yu, C.; Wang, X.; Chen, P.; Xiao, Z.; Han, J.; and Chai, Y. 2024.
\newblock Dynamic mixed-prototype model for incremental deepfake detection.
\newblock In \emph{ACM International Conference on Multimedia}, 8129--8138.

\bibitem[{Wang and Deng(2021)}]{wang2021representative}
Wang, C.; and Deng, W. 2021.
\newblock Representative forgery mining for fake face detection.
\newblock In \emph{IEEE Conference on Computer Vision and Pattern Recognition}, 14923--14932.

\bibitem[{Wang et~al.(2018)Wang, Wang, Zhou, Ji, Gong, Zhou, Li, and Liu}]{wang2018cosface}
Wang, H.; Wang, Y.; Zhou, Z.; Ji, X.; Gong, D.; Zhou, J.; Li, Z.; and Liu, W. 2018.
\newblock CosFace: Large Margin Cosine Loss for Deep Face Recognition.
\newblock In \emph{IEEE Conference on Computer Vision and Pattern Recognition}, 5265--5274.

\bibitem[{Wang et~al.(2022)Wang, Zhang, Yang, Yu, Li, Hong, Zhang, Li, Zhong, and Zhu}]{wang2022memory}
Wang, L.; Zhang, X.; Yang, K.; Yu, L.; Li, C.; Hong, L.; Zhang, S.; Li, Z.; Zhong, Y.; and Zhu, J. 2022.
\newblock Memory replay with data compression for continual learning.
\newblock \emph{arXiv preprint arXiv:2202.06592}.

\bibitem[{Yan, Xie, and He(2021)}]{yan2021dynamically}
Yan, S.; Xie, J.; and He, X. 2021.
\newblock Der: Dynamically expandable representation for class incremental learning.
\newblock In \emph{IEEE Conference on Computer Vision and Pattern Recognition}, 3014--3023.

\bibitem[{Yan et~al.(2024{\natexlab{a}})Yan, Luo, Lyu, Liu, and Wu}]{yan2024transcending}
Yan, Z.; Luo, Y.; Lyu, S.; Liu, Q.; and Wu, B. 2024{\natexlab{a}}.
\newblock Transcending forgery specificity with latent space augmentation for generalizable deepfake detection.
\newblock In \emph{IEEE Conference on Computer Vision and Pattern Recognition}, 8984--8994.

\bibitem[{Yan et~al.(2025{\natexlab{a}})Yan, Wang, Jin, Zhang, Liu, Chen, Yao, Ding, Wu, and Yuan}]{Effort}
Yan, Z.; Wang, J.; Jin, P.; Zhang, K.-Y.; Liu, C.; Chen, S.; Yao, T.; Ding, S.; Wu, B.; and Yuan, L. 2025{\natexlab{a}}.
\newblock Orthogonal Subspace Decomposition for Generalizable AI-Generated Image Detection.
\newblock In \emph{International Conference on Machine Learning}, 70268--70288.

\bibitem[{Yan et~al.(2025{\natexlab{b}})Yan, Wang, Wang, Jin, Zhang, Chen, Yao, Ding, Wu, and Yuan}]{yan2025effort}
Yan, Z.; Wang, J.; Wang, Z.; Jin, P.; Zhang, K.-Y.; Chen, S.; Yao, T.; Ding, S.; Wu, B.; and Yuan, L. 2025{\natexlab{b}}.
\newblock Effort: Efficient orthogonal modeling for generalizable ai-generated image detection.
\newblock In \emph{International Conference on Machine Learning}.

\bibitem[{Yan et~al.(2024{\natexlab{b}})Yan, Yao, Chen, Zhao, Fu, Zhu, Luo, Yuan, Wang, Ding et~al.}]{df40}
Yan, Z.; Yao, T.; Chen, S.; Zhao, Y.; Fu, X.; Zhu, J.; Luo, D.; Yuan, L.; Wang, C.; Ding, S.; et~al. 2024{\natexlab{b}}.
\newblock DF40: Toward Next-Generation Deepfake Detection.
\newblock \emph{arXiv preprint arXiv:2406.13495}.

\bibitem[{Yan et~al.(2023{\natexlab{a}})Yan, Zhang, Fan, and Wu}]{yan2023ucf}
Yan, Z.; Zhang, Y.; Fan, Y.; and Wu, B. 2023{\natexlab{a}}.
\newblock UCF: Uncovering Common Features for Generalizable Deepfake Detection.
\newblock In \emph{IEEE International Conference on Computer Vision}, 22412--22423.

\bibitem[{Yan et~al.(2023{\natexlab{b}})Yan, Zhang, Yuan, Lyu, and Wu}]{deepfakebench}
Yan, Z.; Zhang, Y.; Yuan, X.; Lyu, S.; and Wu, B. 2023{\natexlab{b}}.
\newblock Deepfakebench: A comprehensive benchmark of deepfake detection.
\newblock \emph{arXiv preprint arXiv:2307.01426}.

\bibitem[{Zhang et~al.(2025)Zhang, Zhu, Zhang, Yan, Cheng, Lao, Cai, and Guo}]{gpl}
Zhang, X.; Zhu, P.; Zhang, C.; Yan, Z.; Cheng, J.; Lao, M.; Cai, S.; and Guo, Y. 2025.
\newblock Generalization-Preserved Learning: Closing the Backdoor to Catastrophic Forgetting in Continual Deepfake Detection.
\newblock In \emph{Proceedings of the IEEE/CVF International Conference on Computer Vision}, 3798--3808.

\bibitem[{Zhao et~al.(2021)Zhao, Zhou, Chen, Wei, Zhang, and Yu}]{zhao2021multi}
Zhao, H.; Zhou, W.; Chen, D.; Wei, T.; Zhang, W.; and Yu, N. 2021.
\newblock Multi-attentional deepfake detection.
\newblock In \emph{IEEE Conference on Computer Vision and Pattern Recognition}, 2185--2194.

\bibitem[{Zhou et~al.(2026)Zhou, He, Lin, Fan, Ding, and Li}]{SimplicityAIGI}
Zhou, Y.; He, X.; Lin, K.; Fan, B.; Ding, F.; and Li, B. 2026.
\newblock Simplicity Prevails: The Emergence of Generalizable AIGI Detection in Visual Foundation Models.
\newblock \emph{arXiv preprint arXiv:2602.01738}.

\bibitem[{Zi et~al.(2020)Zi, Chang, Chen, Ma, and Jiang}]{zi2020wilddeepfake}
Zi, B.; Chang, M.; Chen, J.; Ma, X.; and Jiang, Y.-G. 2020.
\newblock Wilddeepfake: A challenging real-world dataset for deepfake detection.
\newblock In \emph{ACM International Conference on Multimedia}, 2382--2390.

\end{thebibliography}

\clearpage
\section*{Appendix}
\setcounter{section}{0}
\section{Related Work} \label{related work}
\subsection{Face Forgery Detection}
Existing generalization-oriented face forgery detection methods typically leverage a limited set of known forged samples to train a generic detector capable of handling unseen forgery types. Early studies primarily enhance discriminative capability by exploiting explicit forgery-related patterns, such as noise artifacts \cite{li2020face}, local region cues \cite{chen2021local,zhao2021multi}, and frequency-domain information \cite{qian2020thinking,guo2020learning,kashiani2025freqdebias}. As forgery techniques become increasingly realistic, explicit artifacts tend to diminish, shifting research focus toward mining implicit forgery patterns. To this end, a variety of learning strategies have been proposed, including contrastive learning \cite{sun2022dual}, identity-aware modeling \cite{huang2023implicit,dong2023implicit}, disentangled representation learning \cite{liang2022exploring,yan2023ucf}, reconstruction-based learning \cite{cao2022end,wang2021representative}, and diverse data augmentation techniques \cite{chen2022self,shiohara2022detecting,yan2024transcending}. More recently, several approaches based on vision Transformers have leveraged vision–language models, such as CLIP \cite{cui2025forensics} and the LoRA-based Effort \cite{yan2025effort}, to further enhance generalization performance. Overall, existing approaches strive to derive forgery-aware representations that generalize beyond known manipulations. Yet, due to the limited coverage of current forgery datasets, detectors trained under a static offline setting struggle to remain effective when confronted with continuously emerging forgery techniques.
\subsection{Incremental Learning for Forgery Detection}
Incremental learning has been widely applied to mitigate catastrophic forgetting, with replay-based strategies dominating the incremental face forgery detection. These methods store or reuse representative samples from previous tasks and employ various enhancements to preserve prior knowledge. For instance, CoReD \cite{kim2021cored} uses knowledge distillation to retain information from past tasks. DFIL \cite{pan2023dfil} prioritizes central and challenging samples to maximize the value of replayed data. HDP \cite{sun2025continual} leverages carefully crafted universal adversarial perturbations for more robust replay. DMP \cite{tian2024dynamic} summarizes previous distributions through mixed prototypes. The most recent method, SUR-LID \cite{cheng2025stacking}, combines sparse uniform replay with a latent-space incremental detector to further consolidate prior knowledge. While effective, all of these approaches rely on explicit access to stored historical data, which imposes non-negligible memory overhead and raises potential privacy concerns.

In continual learning, several studies have explored compressed replay to reduce storage cost and store more historical samples under a fixed memory budget, such as compressing exemplars by removing non-discriminative pixels \cite{luo2023class} or selecting an appropriate compression quality for replay \cite{wang2022memory}. However, existing compressed replay methods are largely developed for generic class-incremental settings and typically operate at the image or pixel level, without explicitly considering the task-specific structure of face forgery detection. In this domain, replay is expected to preserve identity cues, forgery traces, and domain-specific patterns that may exhibit distinct spatial distributions, suggesting that more tailored replay compression strategies still remain to be explored.

\section{Discussion on Real-World Actual Storage Consumption}
Considering that the core objective of this work is to minimize storage consumption in real-world scenarios as much as possible, it is necessary to rigorously analyze the relationship between storage cost and the replay representation from a practical perspective.

Intuitively, one may assume that reducing the effective pixels of a single image by $n\%$ would naturally lead to an equivalent $n\%$ reduction in storage. However, this assumption does not hold in practice. A single sector-shaped region cannot be directly stored as a standalone PNG file. Moreover, simply setting invalid pixels to zero or one does not yield a real reduction in storage cost due to the underlying image encoding mechanism.

To bridge the gap between pixel reduction and actual storage reduction, we adopt a simple yet effective storage strategy. Specifically, instead of storing the masked image directly, we represent each row of the cropped region as a variable-length array containing only the valid pixel values. Combined with the original image resolution, the learnable segment can then be reconstructed online during training, while ensuring that the storage reduction strictly matches the targeted $n\%$.
This design is enabled by the structural property of the ClockMix cropping region, which is always contiguous along \textit{one side} of the image boundary. In other words, for every row, the starting horizontal coordinate must be either 0 or ($1 - \text{len}$), where ($\text{len}$) denotes the valid length of that row. Therefore, we only need to store an additional binary flag per image to indicate whether the segment starts from 0 or from ($1 - \text{len}$). This 1-bit overhead is negligible in practice, yet it allows for lossless reconstruction of the original segment.

In summary, the proposed percentage reduction of the replay region can be translated into a proportional reduction in actual storage cost in practical deployment settings.

% \begin{algorithm*}[h]
% \caption{Replay Storing Algorithm of Infodense}\label{alg:main}
% 		\KwIn{
%         Previous ($t-1$)-Task Dataset $\mathbb{I}^{t-1}$;
%         Previously-trained Detector $f(\cdot,\theta^{t-1})$;
%         Pre-trained score function $\mathbf{s_{\theta}}(\cdot)$.
%         }

%         \For{ $\mathbf{x} \in \mathbb{I}^{t-1}$}{

%             extract image token via encoder from $f(\cdot,\theta^{t-1})$

%             $\{\mathbf{z}_{cls},\mathbf{z}_{(1,1)},...,\mathbf{z}_{(n,n)}\}=\mathit{E}(\mathbf{x})$
            
%             compute forgery similarity score of each patch at location ($i,j$)

%             $\delta_{(i,j)}=\text{Sim}(\mathbf{z}_{(i,j)},\mathbf{z}_{f})$

%             rank score to obtain decisive patches.

%             $\mathbf{P}_D \leftarrow \operatorname{Top}_{10}[\{\mathbf{p}_{(i,j)}\}, \delta_{(i,j)}]$

%             cut out one fixed-size segment that contains the maximum decisive patches 

%             $    \mathbf{s}=\text{Cut}(\mathbf{x}) = \max_{\alpha \in [0, 360-360/n)} \left( \text{Count}_\text{D}(\text{Seg}(v_1,v_2,\alpha)) \right)$

%             store $\mathbf{s}$ into candidate set $\mathbf{S}$
%             }

%         Obtain final InfoDense replay set for ($t-1$)-task

%         $    \mathbf{S}^{t-1} = \operatorname{Top}_{m}[\text{sort}\left(\mathbf{S},\cos(\mathbf{c},f(\mathbf{x}_i))+\text{Count}_\text{D}(\mathbf{s}_{i})\right)]$
        
% 		\KwOut{Final Replay Set $\mathbf{S}^{t-1}$.}  
% \end{algorithm*}

\begin{algorithm*}[h]
\caption{Replay Storing Algorithm of InfoDense}
\label{alg:main}
\begin{algorithmic}[1]
\REQUIRE Previous-task dataset $\mathbb{I}^{t-1}$; trained detector
$f(\cdot,\theta^{t-1})$; forgery text embedding $\mathbf{z}_{f}$;
replay size $m$; weight $\lambda$.
\ENSURE Final replay set $\mathbf{S}^{t-1}$.

\STATE Initialize the candidate set $\mathbf{S} \leftarrow \emptyset$.

\FOR{$\mathbf{x}_i \in \mathbb{I}^{t-1}$}
    \STATE Extract image tokens:
    \[
    \{\mathbf{z}_{cls},\mathbf{z}_{(1,1)},\ldots,
    \mathbf{z}_{(n,n)}\}=\mathit{E}(\mathbf{x}_i).
    \]
    \STATE Compute the forgery-decisive score of each patch:
    \[
    \delta_{(u,v)}
    =
    \operatorname{Sim}(\mathbf{z}_{(u,v)},\mathbf{z}_{f}).
    \]
    \STATE Select the top-$10$ decisive patches:
    \[
    \mathbf{P}_{D}^{i}
    \leftarrow
    \operatorname{Top}_{10}
    \bigl(\{\mathbf{p}_{(u,v)}\},\delta_{(u,v)}\bigr).
    \]
    \STATE Find the segment containing the maximum number of decisive patches:
    \[
    \alpha_i^{*}
    =
    \operatorname*{arg\,max}_{\alpha\in[0,\,360-360/n)}
    \operatorname{Count}_{D}
    \bigl(\operatorname{Seg}(v_1,v_2,\alpha)\bigr).
    \]
    \STATE Extract the corresponding segment:
    \[
    \mathbf{s}_i
    =
    \operatorname{Seg}(\mathbf{x}_i,v_1,v_2,\alpha_i^{*}).
    \]
    \STATE $\mathbf{S}\leftarrow\mathbf{S}\cup
    \{(\mathbf{s}_i,\mathbf{x}_i)\}$.
\ENDFOR

\STATE Compute the feature centroid:
\[
\mathbf{c}
=
\frac{1}{|\mathbb{I}^{t-1}|}
\sum_{\mathbf{x}_i\in\mathbb{I}^{t-1}}
f(\mathbf{x}_i).
\]

\FOR{$(\mathbf{s}_i,\mathbf{x}_i)\in\mathbf{S}$}
    \STATE Compute representativeness and decisive-patch density:
    \[
    R_i=\cos(\mathbf{c},f(\mathbf{x}_i)),
    \qquad
    D_i=\operatorname{Count}_{D}(\mathbf{s}_i).
    \]
\ENDFOR

\STATE Normalize $\{R_i\}$ and $\{D_i\}$ to $[0,1]$, obtaining
$\widetilde{R}_i$ and $\widetilde{D}_i$.

\STATE Compute the selection score:
\[
\operatorname{Score}(\mathbf{s}_i)
=
\lambda\widetilde{R}_i
+
(1-\lambda)\widetilde{D}_i.
\]

\STATE Select the top-$m$ segments:
\[
\mathbf{S}^{t-1}
=
\operatorname{Top}_{m}
\bigl(\mathbf{S},\operatorname{Score}(\mathbf{s}_i)\bigr).
\]

\end{algorithmic}
\end{algorithm*}

\section{Algorithm}
Here, we provide the replay storing details of the proposed InfoDense via Alg.~\ref{alg:main}, which includes cutting and selection.

% 考虑到本文的核心愿景是尽可能的降低真实场景的存储消耗，我们有必要完全基于实际的去分析存储量与replay形式之间的关系。Intuitively, 我们会理所当然的觉得减少单张图n%的有效像素就可以达到同等的减少n%存储的效果。但实际上，单个的扇形区域并不能直接以png的形式被存储。如果直接对无效像素区域进行置0或置1操作，则无法达到实际的降低存储效果。因此，我们配套的用一种简单但有效的存储策略来处理这个问题，来bridge 减少像素至减少存储的gap。具体的说，我们可以直接以不等长数组的形式存储每一行的像素值，这样结合图片的原始分辨率，然后在训练时online的复原得到可学习的segment数据，同时保证存储的减少量为n%.这是因为clockmix的裁剪区域一定是沿着边缘连续的，也就是说所有行数据的开始横坐标必须全是0，或全是1-len，其中len是当前行的长度。因此，我们只需要在该数组中额外的存入一个binary的flag，来指明(1-len) or len，就可以无损的完整复原原始segment。显然，这个1bit的额外flag开销是可以几乎忽略不计的。综上所述，我们对于replay region的百分比减少，是可以等价的转换为实际应用中对于存储的减少的。

\section{Detailed Implementation Settings}

\subsection{Preprocessing}

\paragraph{Data Preprocessing.}
Following the standard DeepFakeBench~\cite{deepfakebench} protocol, all video frames from all datasets undergo face detection, extraction, and alignment before being resized to $224 \times 224$. For input normalization, we adopt a mean of $[0.4815, 0.4578, 0.4082]$ and a standard deviation of $[0.2686, 0.2613, 0.2758]$ for the three RGB channels, which are adopted by the official CLIP-series models~\cite{clip}. During training, 8 frames are uniformly sampled from each video, while 32 frames are sampled during testing to ensure more stable and reliable performance evaluation, as well as alignment with baselines.
% 按照 DeepFakeBench 的标准协议，所有视频帧通过人脸检测与对齐进行预处理，之后将人脸图像缩放至 $224 \times 224$。归一化采用均值 $$ 和标准差 $$。训练阶段每个视频采样 4 帧，测试阶段采样 32 帧，以增强评估的稳定性。

\paragraph{Data Augmentation.}
To enhance the generalization ability of the detector, we employ a comprehensive data augmentation pipeline on the current task data using the \texttt{Albumentations~\cite{buslaev2020albumentations}} library. The augmentation operations and their corresponding application probabilities are detailed as follows:

% 为了提升检测器的泛化能力，我们使用 \texttt{Albumentations} 库对当前任务的数据应用严格的数据增强流程。具体的增强操作及其概率如下：

\begin{itemize}
\item \textbf{Spatial Transformations:} Horizontal Flip ($p=0.5$), Rotation within $\pm 10^\circ$ ($p=0.5$), and isotropic resizing.
\item \textbf{Pixel-level Transformations:} Gaussian Blur with a kernel size in the range of $[3, 7]$ ($p=0.5$).
\item \textbf{Compression Artifacts:} JPEG compression with a quality range of 40--100, applied with $p=0.5$.
\item \textbf{Color Perturbations:} One of Random Brightness/Contrast (limit 0.1), FancyPCA, or HueSaturationValue, selected and applied with probability $p=0.5$.
\end{itemize}

It is worth noting that while the current task data are augmented as described above, the generated replay samples are kept with standard normalization only so as to preserve their intrinsic generative distribution characteristics.

\subsection{Batch Composition and Fusion Details}

To ensure strict class balance during the incremental learning process, we employ a paired sampling strategy for the current-task data. Within each training step, the base batch is explicitly enforced to contain an equal quantity of intact pristine and manipulated images, denoted as $B_r$ real samples and $B_f$ fake samples. To promote model desensitization to artificial fusion boundaries, we first construct intra-task mixed samples. Given a spatial compression rate $\gamma$, which represents the reciprocal of the retained sector angle, we synthesize $\lfloor B_r / \gamma \rfloor$ intra-task real-fused samples alongside $\lfloor B_f / \gamma \rfloor$ intra-task fake-fused samples directly from the current batch. Furthermore, to consolidate historical knowledge, we execute inter-task fusion governed by a replay batch parameter $B_{rep}$. Specifically, we sample localized fragments from the memory buffer and fuse them onto current-task images, yielding three distinct cross-task combinations to prevent class bias: $B_{rep}$ samples combining previous real with current real, $B_{rep}$ samples combining previous fake with current fake, and $B_{rep}$ samples combining previous fake with current real. All aforementioned fused variants, together with the original intact samples, constitute the final comprehensive mini-batch for detector optimization.In our implementation, we empirically set $B_r=8$ and $B_{rep}=4$.

% \subsection{批次构成与融合细节}
% 为了在增量学习过程中确保严格的类别平衡，我们对当前任务数据采用了配对采样策略。在每个训练步骤中，基础批次被明确强制包含等量的完整真实和伪造图像，分别记为 B_r 个真实样本和 B_f 个伪造样本。为了促进模型对人为融合边界的脱敏，我们首先构建任务内混合样本。给定一个空间压缩率 \gamma（代表保留扇形角度的倒数），我们直接从当前批次中合成 \lfloor B_r / \gamma \rfloor 个任务内真实融合样本，以及 \lfloor B_f / \gamma \rfloor 个任务内伪造融合样本。此外，为了巩固历史知识，我们执行由回放批次参数 B_{rep} 控制的跨任务融合。具体而言，我们从内存缓冲区中采样局部片段并将其融合到当前任务图像上，产生三种不同的跨任务组合以防止类别偏差：B_{rep} 个结合了先前真实与当前真实的样本、B_{rep} 个结合了先前伪造与当前伪造的样本，以及 B_{rep} 个结合了先前伪造与当前真实的样本。所有上述融合变体连同原始完整样本，共同构成了用于检测器优化的最终综合微批次。我们设置 Br=8, Brep=4.

\section{Further Analysis and Ablations}

\begin{figure*}[t]
\centering
\begin{minipage}{0.58\textwidth}
  \centering
  \includegraphics[width=\linewidth]{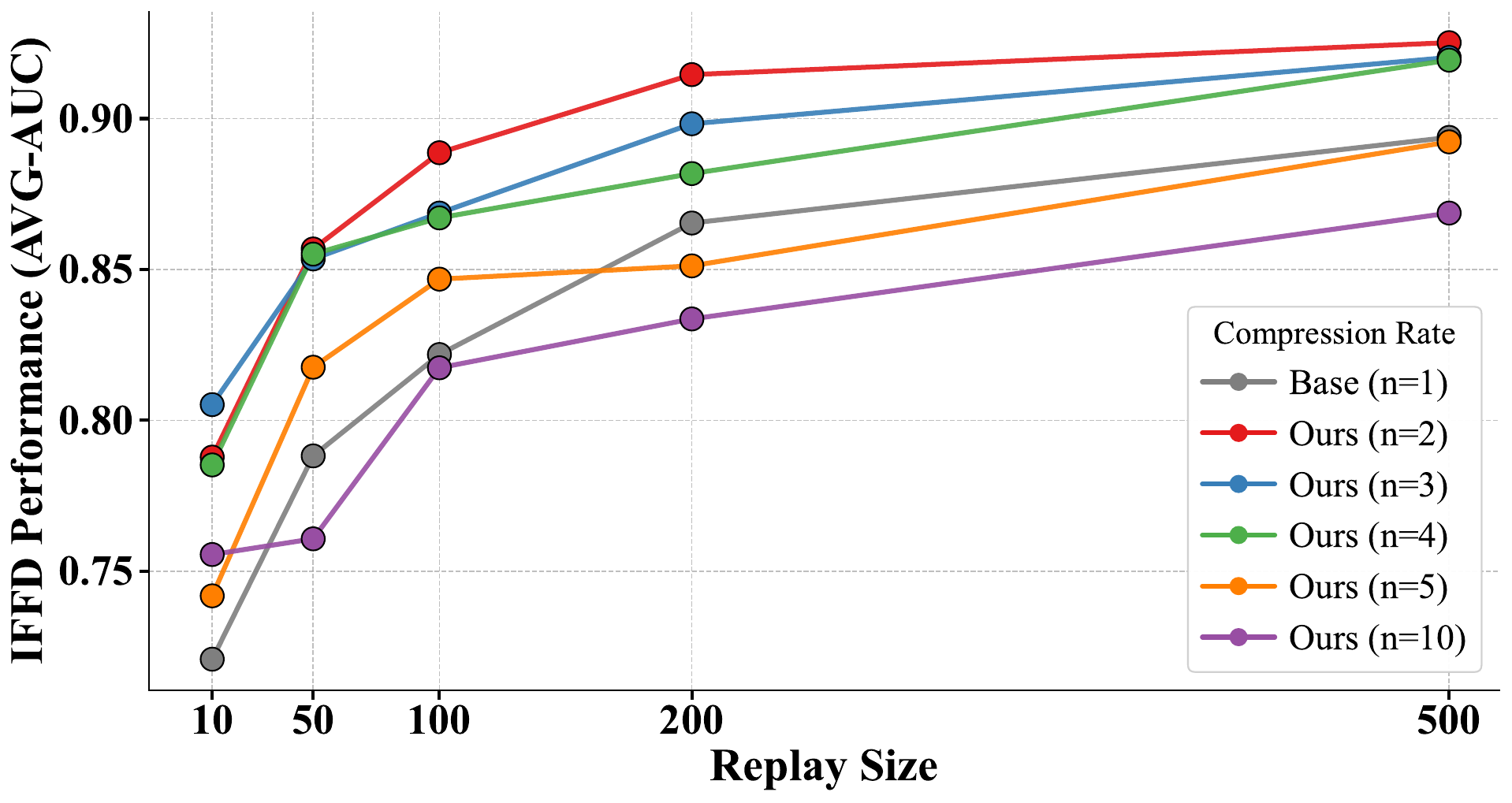}
  \caption{Performance comparison strictly aligned by the replay buffer size. }
  % 图1 Caption：严格按照回放缓冲区大小对齐的性能比较。
  \label{fig:size}
\end{minipage}
\hfill
\begin{minipage}{0.40\textwidth}
  \centering
  \includegraphics[width=\linewidth]{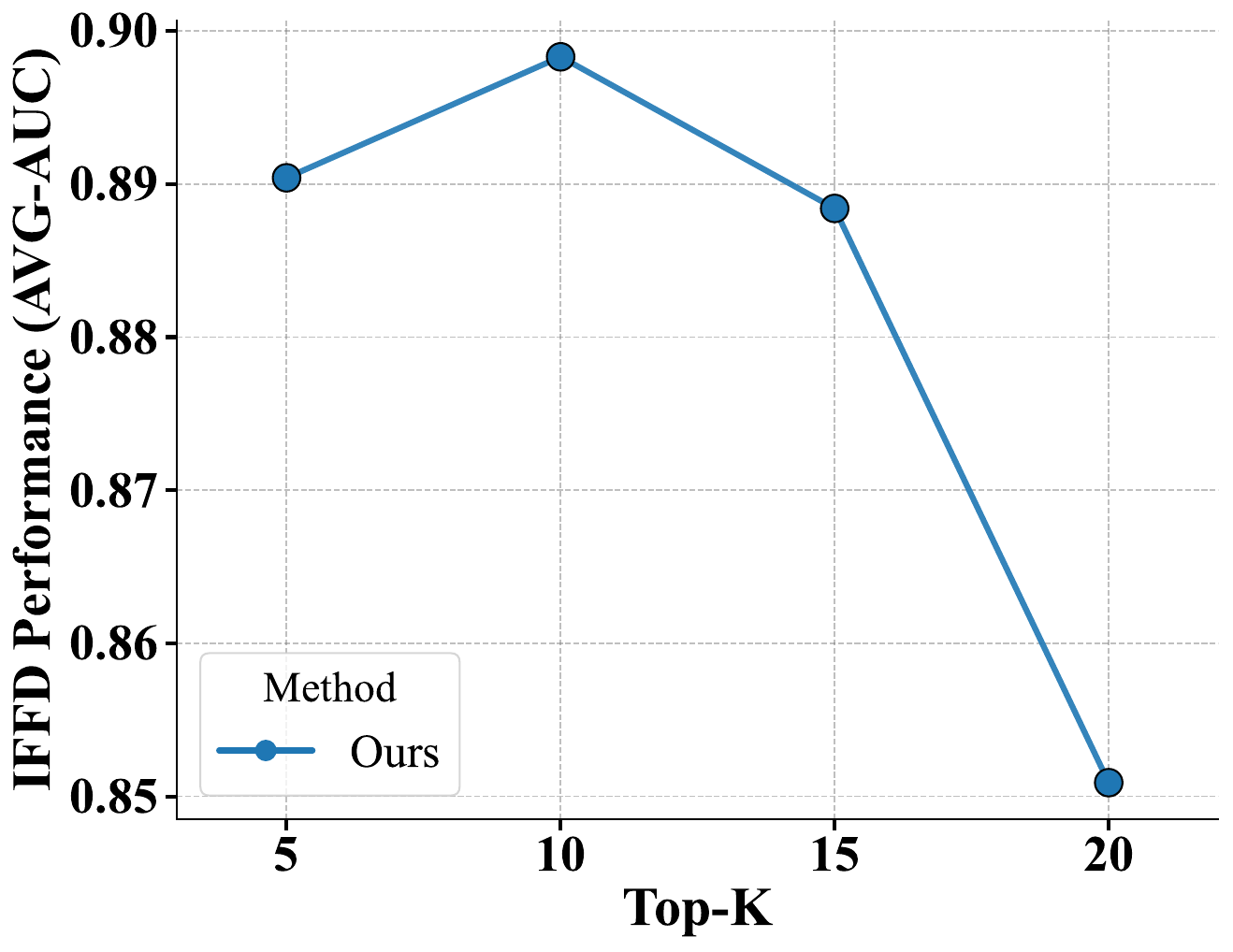}
  \caption{Sensitivity analysis on the number of decisive patches $K$.}
  % 图2 Caption：对用于提取引导的决定性 Patch 数量 K 的超参数敏感性分析。
  \label{fig:topk}
\end{minipage}
\end{figure*}

\subsection{Performance Evaluation under Aligned Buffer Sizes}
To provide a complementary perspective to the storage-aligned Pareto analysis in Fig.~\ref{fig:storage}, we further evaluate the performance by strictly aligning the replay buffer size (\textit{i.e.}, the absolute numerical quantity of stored samples). As shown in Fig.~\ref{fig:size}, when utilizing moderate compression rates ($n \in \{2, 3, 4\}$), the InfoDense strategy consistently matches or surpasses the full-image baseline across varying buffer capacities. This phenomenon reveals that full images inherently contain task-irrelevant background noise and dataset-specific biases that can hinder incremental learning. By proactively filtering out these redundancies, moderate density-aware extraction provides a purer, more concentrated supervision signal. Conversely, excessively aggressive compression ($n=5$ or $10$) causes performance to drop below the baseline, indicating that extreme spatial reduction inevitably discards essential forgery artifacts alongside the background. Therefore, appropriate regional extraction acts as a powerful regularization mechanism that mitigates domain-specific redundancy while retaining core discriminative clues, thereby yielding strong generalization even without the advantage of an expanded sample capacity.

% \subsection{对齐缓冲区大小下的性能评估}
% 为了对按存储对齐的 Pareto 分析提供补充视角，我们通过严格对齐回放缓冲区大小（即存储样本的绝对数值量）来进一步评估性能。如图~\ref{size} 所示，当使用适中的压缩率（$n \in \{2, 3, 4\}$）时，InfoDense 策略在不同的缓冲区容量下始终达到或超越了全图基线。这一现象揭示了全图本身包含与任务无关的背景噪声和特定数据集的偏差，这会阻碍增量学习。通过主动过滤掉这些冗余，适度的密度感知提取提供了更纯粹、更集中的监督信号。相反，过度激进的压缩（n=5 或 10）会导致性能降至基线以下，这表明极端的空间缩减不可避免地会在丢弃背景的同时丢弃基本的伪造伪影。因此，适当的区域提取作为一种强大的正则化机制——在去除特定领域噪声的同时保留了核心判别线索——从而即使在没有样本容量扩张优势的情况下，也能实现强大的泛化能力。

\subsection{Sensitivity Analysis of Decisive Patches}

We conduct a sensitivity analysis on the number of decisive patches $K$ used during density-aware extraction. As shown in Fig.~\ref{fig:topk}, the incremental detection performance reaches its best result at $K=10$ and degrades when $K$ is either too small or too large. This trend supports our facial redundancy hypothesis that a limited number of decisive patches is sufficient to preserve the forgery-related information required for incremental learning. Specifically, when $K$ is too small (\textit{e.g.}, $K=5$), the selected region may fail to cover some fine-grained forgery artifacts. In contrast, a larger $K$ (\textit{e.g.}, $K=20$) introduces more identity-related and task-irrelevant facial regions, reducing the information density of the extracted segment. Overall, $K=10$ provides the best trade-off between preserving discriminative forgery cues and avoiding unnecessary spatial redundancy.

\begin{figure*}[t]
    \centering
    \includegraphics[width=1\linewidth]{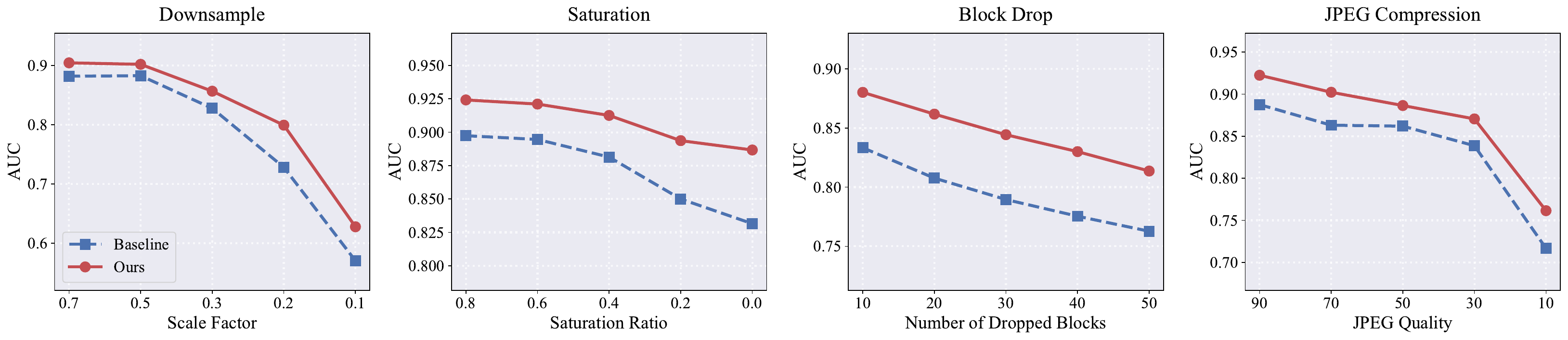}
    \caption{Robustness evaluation against test-time common image degradations across five progressive severity levels.}
    \label{fig:robustness}
\vspace{-0.3cm}
\end{figure*}
% \subsection{决定性 Patch 的敏感性分析}
% 我们对密度感知提取阶段使用的决定性 Patch 数量 K 进行了超参数敏感性分析。如图~\ref{topk} 所示，增量检测性能在 K=10 时达到峰值，并在 K 设置过小或过大时出现明显下降。这一趋势从经验上验证了我们的人脸冗余假设。如果 K 受到过度限制（例如 K=5），提取的扇区就有遗漏关键、细粒度伪造伪影的风险。相反，过大的 K（例如 K=20）会迫使指定扇区扩展到冗余的、与身份相关的平滑皮肤区域，从而稀释了整体信息密度并重新引入了与任务无关的噪声。因此，K=10 实现了最佳平衡，完美地封装了最集中的判别性伪影，而没有捕获不必要的空间冗余。

\section{Qualitative Visualization}
\begin{figure}[htbp]
    \centering
    \includegraphics[width=\linewidth]{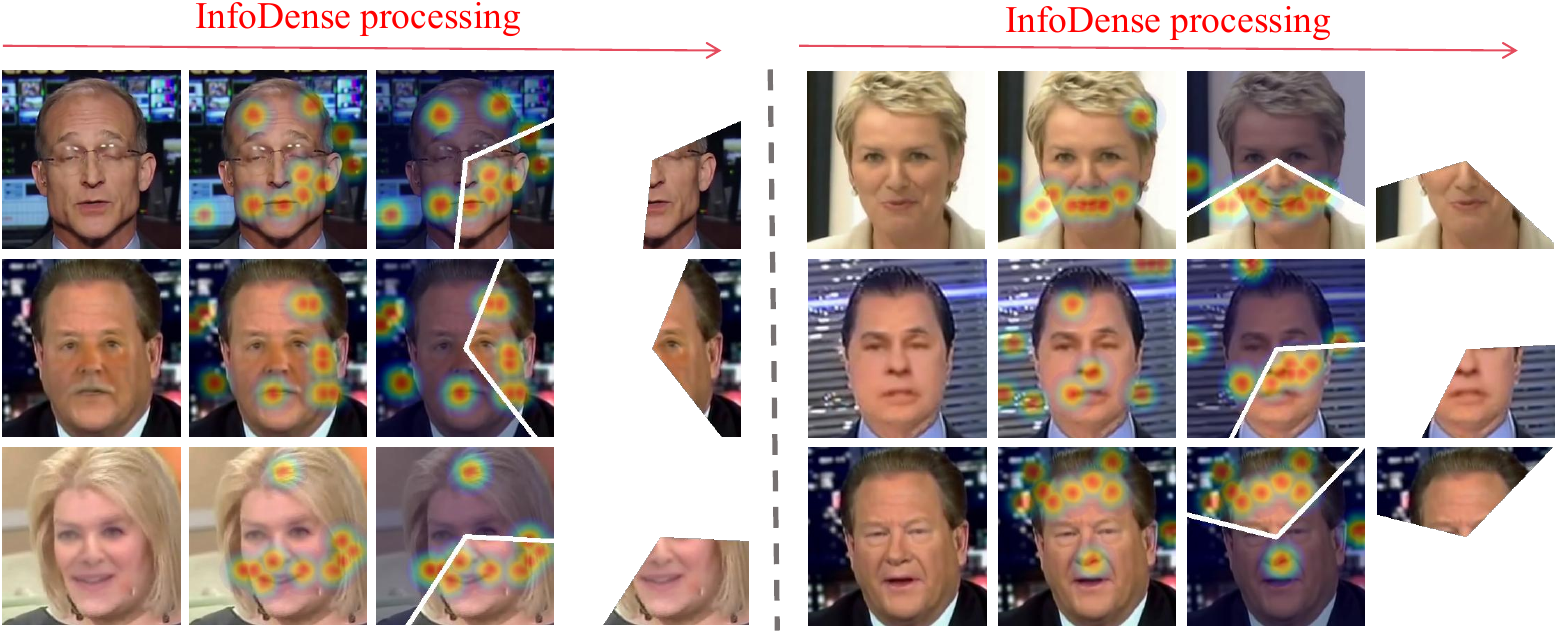}
    \caption{Visualization of InfoDense extraction. From left to right: (a) Original face. (b) Top-$K$ decisive patches. (c) Cropping region. (d) Final InfoDense replay segment.}
    \label{fig:sample}
\end{figure}

To intuitively demonstrate the density-aware extraction process, Figure~\ref{fig:sample} provides a step-by-step visualization of how InfoDense progressively compresses each input face. Our method accurately localizes fine-grained forgery artifacts and removes task-irrelevant background regions, thereby achieving substantial storage compression while preserving the essential discriminative clues required for effective replay.
%为了直观展示我们的密度感知提取，图~\ref{fig:sample} 可视化了逐步压缩的过程。我们的方法准确地定位了细粒度的伪造伪影并丢弃了与任务无关的背景，在实现极端存储压缩的同时，保留了用于演练的基本判别线索。图例：InfoDense 提取的定性可视化。从左至右：(a) 原始人脸。(b) 前 $K$ 个决定性 Patch(c)切割示意图 。(d) 最终的 InfoDense 回放片段。

% ================= Rebuttal-added Appendix Experiments =================

\providecommand{\ours}{\textbf{Ours}}

\section{Additional Experimental Results}
\label{sec:additional_experimental_results}
\subsection{Robustness against Image Degradations}
To verify that the proposed regional decisive replay mechanism does not overfit to high-quality pristine data, we evaluate its robustness against common real-world image degradations. During the inference phase, we independently apply four types of perturbations, including downsampling, saturation alteration, block-wise drop, and JPEG compression. Each degradation is designed with five progressive severity levels to comprehensively assess the performance decay trend, with average AUC as the evaluation metric. As shown in Fig.~\ref{fig:robustness}, InfoDense consistently exhibits superior resilience against varying degrees of corruption compared to the baseline. This demonstrates that by selectively replaying information-dense regions, the model reduces reliance on fragile artifacts and successfully learns robust forgery traces.
\subsection{Prompt and CLIP Intrinsic Knowledge for Forgery Awareness}\label{sec:supp-clip}

The decisive-patch score is computed using a forgery-trained CLIP, rather than an untouched zero-shot CLIP. Namely, the score is not used in the first task, but only in subsequent tasks where \textit{replay} is required, by which time the model has \textit{already} learned forgery-awareness from previous tasks.

Moreover, the pretrained CLIP representation already contains forgery-discriminative information. Prior works~\cite{ForAda,SimplicityAIGI,Effort} and our Table~\ref{tab:clip_linear_ffpp} support that a frozen pretrained CLIP backbone with only a trained LoRA and linear classifier can achieve strong detection, suggesting its intrinsic non-trivial forgery separability. Therefore, we adopt a frozen backbone with an Adapter that can preserve the overall representation structure and text-image alignment ability of CLIP, with ``this image is [Learn-Token] \{real, fake\}'' as the prompts.

The score is guiding as a \textit{better-than-random} prior for selecting more informative regions, not as a precise forgery mask estimator. Hence, it is sufficient for the score to prioritize artifact-rich regions; occasional sub-optimal selections do not fundamentally undermine our method. Meanwhile, Fig.~\ref{fig:sample} shows the effectiveness of leveraging CLIP-similarity to localize forgery regions.

\begin{table}[t]
\centering
\caption{\small Frozen CLIP for intrinsic forgery capability.}
\label{tab:clip_linear_ffpp}
\small
\setlength{\tabcolsep}{6pt}
\renewcommand{\arraystretch}{0.95}
\begin{tabular}{lccc}
\toprule
Method & FF++ & CDFv2 & Avg. \\
\midrule
CLIP+Linear & 87.33 & 94.54 & 90.94 \\
\bottomrule
\end{tabular}
\end{table}

\subsection{Weight Ablation and Fragment Number Analysis}

In practice, we normalize both terms to $[0,1]$ before combination. Table~\ref{tab:weight_ablation} reports the results with different values of $\lambda$ in Eq.~(4). The results show that jointly considering representativeness and decisive-patch density is more effective than relying on either term alone, with $\lambda=0.5$ used as our default setting.

We also analyze whether selecting multiple fragments from one image is beneficial. As shown in the lower half of Table~\ref{tab:weight_ablation}, selecting two or three fragments from the same image degrades performance, possibly because excessive fragmentation disrupts local forgery structures.

\begin{table*}[htbp]
\centering
\caption{\small Upper half: ablation on $\lambda$ in Eq.~(4), where $\lambda$ weights representativeness and $1-\lambda$ weights decisive-patch density. Lower half: ``Frag=2'' denotes selecting two different fragments from each image.}
\label{tab:weight_ablation}
\small
\setlength{\tabcolsep}{7pt}
\renewcommand{\arraystretch}{0.85}
\begin{tabular}{lcccccccccc}
\toprule
\multirow{2}{*}{Setting}
& \multicolumn{5}{c}{Incremental Dataset}
& \multicolumn{5}{c}{Cross-dataset} \\
\cmidrule(lr){2-6} \cmidrule(lr){7-11}
& FF++ & DFDCP & CDF & AA $\uparrow$ & AF $\downarrow$
& SDv21 & UADFV & WDF & DFD & Avg. \\
\midrule
$\lambda=0$     & 82.29 & 88.51 & 96.97 & 89.25 & 7.85
& 93.06 & 92.46 & 72.76 & 82.49 & 85.19 \\
$\lambda=1/3$   & 83.01 & 87.65 & 97.59 & 89.41 & 7.67
& 82.01 & \textbf{96.60} & 75.33 & 86.27 & 85.05 \\
$\lambda=2/3$   & 78.10 & 82.89 & 94.56 & 85.18 & 11.67
& 76.25 & 83.68 & 71.09 & 78.98 & 77.50 \\
\midrule
Frag=2 & 82.25 & \textbf{89.81} & 95.97 & 89.34 & 7.43
& 92.60 & 95.05 & 71.01 & 84.55 & 85.80 \\
Frag=3 & 80.71 & 85.19 & 96.96 & 87.62 & 8.86
& 91.19 & 91.65 & 73.10 & 78.99 & 83.73 \\
\midrule
\ours~($\lambda=0.5$)
& \textbf{84.11} & 89.41 & \textbf{97.75} & \textbf{90.42} & \textbf{6.03}
& \textbf{97.88} & 94.71 & \textbf{79.38} & \textbf{88.13} & \textbf{90.03} \\
\bottomrule
\end{tabular}
\end{table*}

\subsection{Analysis of the Fan-shaped Region Strategy}

Since artifact characters in one sample are usually similar, InfoDense does not aim to preserve all forgery regions within a sample. Instead, it retains the most information-dense intact region while discarding less informative regions to improve replay efficiency. The fan-shape covers face and context simultaneously, including facial landmarks, boundaries, blending clues, and background. Table~\ref{tab:fanshape} shows that it outperforms alternative region-selection strategies.

\begin{table}[t]
\centering
\caption{\small Comparison of different region construction strategies. 
LM denotes facial landmark guidance.}
\label{tab:fanshape}
\small
\setlength{\tabcolsep}{8pt}
\renewcommand{\arraystretch}{1.05}
\begin{tabular*}{0.8\linewidth}{@{\extracolsep{\fill}}lccc@{}}
\toprule
\multirow{2}{*}{Method}
& \multicolumn{3}{c}{Cross-dataset AUC (\%)} \\
\cmidrule(lr){2-4}
& DFD & CDF-v2 & Avg. \\
\midrule
Mosaic     & 71.01 & 59.39 & 65.20 \\
CutMix     & 83.99 & 78.62 & 81.31 \\
Mixup      & 81.67 & 75.75 & 78.71 \\
LM-Guided  & 84.09 & 80.81 & 82.45 \\
\midrule
Fan-shaped (Ours)
& \textbf{87.65}
& \textbf{83.94}
& \textbf{85.80} \\
\bottomrule
\end{tabular*}
\end{table}

\subsection{Quantitative Definition of Information Density}

Theoretically, InfoDense can be regarded as semantic-guided replay distillation, aiming to maximize discriminative information per storage unit rather than preserve raw pixels. We define \textit{information density} as
\begin{equation}
D(s)=\frac{1}{|s|}\sum_{p\in P_D}\mathbb{I}(p\in s),
\end{equation}
% where $P_D$ denotes forgery-decisive patches. The regional selection ablation quantitatively shows that InfoDense outperforms random, center-based, and center+hard segment selection.

where $P_D$ denotes forgery-decisive patches. The regional selection ablation quantitatively shows that InfoDense outperforms random, center-based, and center+hard segment selection, demonstrating the effectiveness of jointly considering representativeness and forgery-information density.

\subsection{Comparison under Prior Incremental Protocols}

We further report the protocol aligned with prior works, including DevFD, with lower-triangular stage-wise metrics. This protocol uses the sequence FF++, DFDCP, DFD, and CDF2. The results show that our method remains competitive under prior protocols while still achieving strong final-stage average performance overall.

\begin{table}[H]
\centering
\caption{\small Protocol aligned with prior works, including DevFD, with lower-triangular stage-wise matrix.}
\label{tab:dmp_devfd_ours}
\small
\setlength{\tabcolsep}{3.3pt}
\renewcommand{\arraystretch}{0.85}
% \resizebox{\textwidth}{!}{
\begin{tabular}{l|c|cccc|cc}
\toprule
% \multirow{2}{*}{Method}
% & \multirow{2}{*}{Dataset}
% & \multicolumn{4}{c|}{Acc(%) $\uparrow$}
% & \multirow{2}{*}{Avg. $\uparrow$}
% & \multirow{2}{*}{AF $\downarrow$} \\
\multirow{2}{*}{Method}
& \multirow{2}{*}{Dataset}
& \multicolumn{4}{c|}{Acc(\%) $\uparrow$}
& \multirow{2}{*}{Avg. $\uparrow$}
& \multirow{2}{*}{AF $\downarrow$} \\
\cmidrule(lr){3-6}
& & FF++ & DFDCP & DFD & CDF2 & & \\
\midrule
\multirow{4}{*}{DMP}
& FF++  & \underline{95.96} & -     & -     & -     & \underline{95.96} & -    \\
& DFDCP & \underline{92.71} & 89.72 & -     & -     & 91.22 & 3.25 \\
& DFD   & \textbf{92.64} & 86.09 & 94.84 & -     & 91.19 & \underline{3.48} \\
& CDF2  & \textbf{91.61} & 84.86 & 91.81 & \underline{91.67} & \underline{89.99} & 4.08 \\
\midrule
\multirow{4}{*}{DevFD}
& FF++  & \textbf{98.41} & -     & -     & -     & \textbf{98.41} & -    \\
& DFDCP & \textbf{97.06} & \underline{89.90} & -     & -     & \textbf{93.48} & \textbf{1.35} \\
& DFD   & \underline{92.44} & \textbf{89.07} & \textbf{97.91} & -     & \textbf{93.14} & \textbf{3.40} \\
& CDF2  & \underline{90.71} & \textbf{90.31} & \underline{93.12} & 85.15 & 89.82 & \underline{4.03} \\
\midrule
\multirow{4}{*}{Ours}
& FF++  & 95.02 & -     & -     & -     & 95.02 & -    \\
& DFDCP & 92.12 & \textbf{90.49} & -     & -     & \underline{91.31} & \underline{2.90} \\
& DFD   & 88.99 & \underline{88.49} & \underline{97.82} & -     & \underline{91.77} & 4.02 \\
& CDF2  & 87.73 & \underline{89.96} & \textbf{95.27} & \textbf{96.05} & \textbf{92.61} & \textbf{3.46} \\
\bottomrule
\end{tabular}
% }
\end{table}

\subsection{Additional AA/AF Metrics}

\begin{figure}[htbp]
\centering
\includegraphics[width=0.75\linewidth]{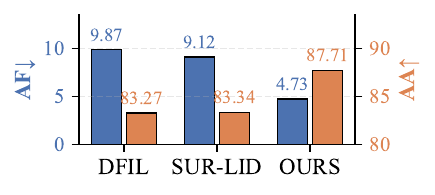}
\caption{\small AA/AF results.}
\label{fig:aaaf}
\end{figure}

We further provide AA/AF results and lower-triangular stage-wise metrics. As shown in Figure~\ref{fig:aaaf} and Table~\ref{tab:dmp_devfd_ours}, our method achieves stability under incremental learning.

\subsection{Computation Overhead}

InfoDense is efficient and GPU-free, adding only \textbf{0.32h} training overhead, corresponding to \textbf{2.70\%}. Our FLOPs remain unchanged since the base model is unmodified.

\subsection{Privacy Analysis of Fused Training Samples}\label{sec:supp-privacy}

Although InfoDense stores only fragmented facial regions, InfoFuse reconstructs
complete training samples by combining historical fragments with current-task
images. We therefore further examine whether these fused samples reveal the
identities of their source faces. Specifically, we use ArcFace and CosFace to
retrieve the corresponding original identities and report Top-1/Top-3 retrieval
recall and identity similarity (IDSim). Lower values indicate less identity
information leakage. As shown in Table~\ref{tab:face_recall}, all retrieval
recalls remain below $0.02\%$, while the identity similarities are close to
zero under both recognition models. These results demonstrate that the fused training samples do not preserve identifiable facial representations and thus provide strong protection against identity leakage.

\begin{table}[H]
\centering
\caption{\small Identity leakage analysis of fused training samples.
Lower values indicate better privacy preservation.}
\label{tab:face_recall}
\small
\setlength{\tabcolsep}{9pt}
\renewcommand{\arraystretch}{1.05}
\begin{tabular}{lcc}
\toprule
Metric $\downarrow$ & ArcFace & CosFace \\
\midrule
Recall@1 (\%) & 0.01 & 0.01 \\
Recall@3 (\%) & 0.02 & 0.01 \\
IDSim         & 0.0071 & 0.0095 \\
\bottomrule
\end{tabular}
\end{table}


\end{document}